\documentclass[letterpaper]{article} 
\usepackage{aaai25}  
\usepackage{times}  
\usepackage{helvet}  
\usepackage{courier}  
\usepackage[hyphens]{url}  
\usepackage{graphicx} 
\usepackage{natbib}  
\usepackage{caption} 
\usepackage[ruled,noline,nofillcomment]{algorithm2e}

\usepackage{float}
\usepackage{subcaption}

\usepackage{multirow}

\usepackage{multicol}

\usepackage{color, colortbl}
\usepackage{xcolor}

\usepackage{pifont} 
\usepackage{bbding} 
\newcommand{\ua}[1]{#1$\uparrow$}               
\newcommand{\da}[1]{#1$\downarrow$}             

\usepackage{ragged2e}
\usepackage{blindtext}
\usepackage{amsmath}
\usepackage{mathtools}
\usepackage{booktabs}
\usepackage{amssymb}
\usepackage{makecell}

\usepackage{newfloat}
\usepackage{listings}
\DeclareCaptionStyle{ruled}{labelfont=normalfont,labelsep=colon,strut=off} 
\floatstyle{ruled}
\newfloat{listing}{tb}{lst}{}
\floatname{listing}{Listing}
\title{UniDemoiré: Towards Universal Image Demoiréing with Data Generation and Synthesis}
\author{
    Zemin Yang\textsuperscript{\rm 1,}\equalcontrib,
    Yujing Sun\textsuperscript{\rm 2,}\equalcontrib,
    Xidong Peng\textsuperscript{\rm 1},
    Siu Ming Yiu\textsuperscript{\rm 2},
    Yuexin Ma\textsuperscript{\rm 1,}\thanks{Corresponding author.}
}
\affiliations{
    \textsuperscript{\rm 1}ShanghaiTech University \\
    \textsuperscript{\rm 2}The University of Hong Kong\\

    \{csyangzm, mayuexin\}@shanghaitech.edu.cn, \{yjsun, smyiu\}@cs.hku.uk
}

\begin{document}

\maketitle

\begin{abstract}
Image demoiréing poses one of the most formidable challenges in image restoration, primarily due to the unpredictable and anisotropic nature of moiré patterns. Limited by the quantity and diversity of training data, current methods tend to overfit to a single moiré domain, resulting in performance degradation for new domains and restricting their robustness in real-world applications. In this paper, we propose a universal image demoiréing solution, \textbf{UniDemoiré}, which has superior generalization capability. Notably, we propose innovative and effective data generation and synthesis methods that can automatically provide vast high-quality moiré images to train a universal demoiréing model. Our extensive experiments demonstrate the cutting-edge performance and broad potential of our approach for generalized image demoiréing. 
\end{abstract}

\begin{links} 
    \link{Code}{https://github.com/4DVLab/UniDemoire} 
\end{links}

\section{Introduction}
\label{sec:intro}
Digital screens have become essential devices for displaying information in our daily work and life.
However, images captured from screens frequently suffer from frustrating moiré patterns, significantly degrading image quality and hindering content extraction. 
Therefore, it becomes crucial to effectively remove such moiré artifacts to help users obtain high-quality images from their digital imaging devices and to support industries in maintaining high-standard product visual presentation and digital archiving.
However, moiré patterns are characterized as anisotropic and multi-scale, as well as involving considerable shape variations and color distortions~\cite{amidror2009theory}. Such traits are seldom seen in other types of artifacts, like noise, rain streaks, fog, blurring, etc., posing a significant challenge for even the most advanced image restoration methods~\cite{luo2023refusion,zhu2023denoising,fei2023generative}. 

Hence, many methods have been proposed to tackle the problem of demoiréing in recent years~\cite{sun2018moire,liu2020wavelet,luo2020deep,he2019mop,he2020fhde,wang2023coarse,yue2022recaptured,yu2022towards}. Nevertheless, the effectiveness of such supervised methods heavily depends on the volume of training data, consisting of pairs of moiré images and their clean counterparts. As we know, collecting such data is a daunting task and it requires precise calibration between natural images and moiré patterns. The limitations of the data lead to the limitations of the methods, resulting in poor generalization of the network model, which performs poorly on the data containing new moiré patterns or new natural images. In order to expand the quantity and diversity of the training data in a convenient way, some methods have started to explore the synthesis of moiré patterns. LCDMoiré~\cite{yuan2019aim} deigns handcraft mathematical models. However, it could not represent complex features of moiré patterns and leads to a substantial discrepancy between the synthetic data and actual moiré images. To enhance realism, recent studies~\cite{cyclic, undem} extract moiré patterns from existing real images and combine them with clean images for data synthesis. Nevertheless, these methods do not escape from the moiré domains of the existing training data, bringing limited performance improvement on new moiré domains. To develop a universal model for image demoiréing with greater generalization capability and practicality, two critical challenges emerge: \textit{how to generate a vast amount of diverse data, and how to ensure the authenticity of the data?}

\begin{figure*}[ht]
  \centering   
  \includegraphics[width=1.0\linewidth]{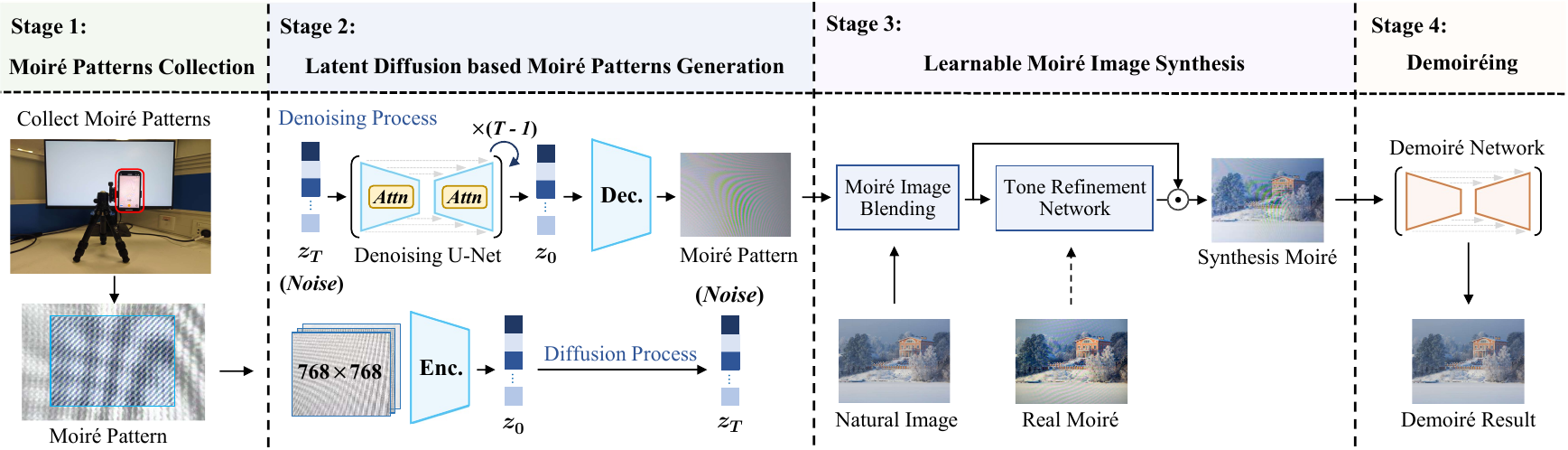}
  \caption{The workflow of our proposed UniDemoiré.}
  \label{fig:pipeline}
\end{figure*}

To address the above challenges, we propose a universal image demoiréing solution, \textbf{UniDemoiré}, capable of generating a vast amount of realistic-looking training data to enhance the generalization capabilities of the image demoiréing model, as Figure~\ref{fig:pipeline} shows. 
First, inspired by the fact that the moiré pattern is unrelated to the content of the image, we introduce a novel, large-scale \textbf{Moiré Pattern Dataset} by capturing moiré patterns against a plain white background. Unlike previous moiré datasets that capture nature images with moiré, our pure moiré patterns can be applied to arbitrary nature images to scale up the data domain automatically. Moreover, our dataset does not need calibrations between the moiré image and the clean image, which can avoid the effect of calibration errors and facilitate the learning process of the model. In particular, our dataset introduces more pattern diversity by considering various previously overlooked factors~\cite{yang2023doing}, including zooming rate, CMOS technology, pixel size, and panel types. Second, building on this real-captured moiré pattern dataset, we propose a diffusion model-based \textbf{Moiré Pattern Generation} method to further increase the diversity of moiré patterns. Specifically, we implement a multi-scale cropping strategy to accommodate different input image sizes and an effective data filtering strategy to ensure the quality of training data for the diffusion model. 
Third, we propose a \textbf{Moiré Image Synthesis} method to create a sufficient amount of diverse and realistic-looking moiré images by blending the generated moiré patterns with clean natural images. In particular, to improve the authenticity of our synthesized data, we develop an effective learnable network and three effective losses to closely mimic the real captured moiré images in terms of color and brightness. Finally, our synthesized abundant moiré images serve to train an \textbf{Image Demoiréing Model} that achieves superior performance and promising generalization capabilities for zero-shot image demoiréing and cross-domain evaluations.

Our contribution can be summarized as follows:
\begin{itemize}
    \item We propose a universal demoiréing solution, which substantially enlarges the knowledge domain and improves the generalization capability of demoiréing models.
    \item We collect a large-scale and high-resolution moiré pattern dataset and develop an effective moiré pattern generator to further increase the diversity of moiré patterns.
    \item We present a novel moiré image synthesis approach, providing a large amount of realistic-looking and high-quality moiré image samples, facilitating the training of a universal image demoiréing model.    
\end{itemize}

\section{Related Work}
\label{sec:related-work} 

\subsection{Image Restoration and Demoiréing}
The inherent complexity of moiré patterns presents a unique challenge compared to other artifacts such as noise~\cite{xing2021end}, haze~\cite{li2021dehazeflow}, blur~\cite{lee2021iterative}, multiple artifacts in one go~\cite{luo2023refusion,zhu2023denoising,fei2023generative, zhang2023all}, etc. Consequently, these methods may not effectively solve the moiré issue. Current mainstream methods for image demoreing are learning based~\cite{sun2018moire,liu2020wavelet,luo2020deep,he2019mop,he2020fhde,niu2023progressive,wang2023coarse, yue2022recaptured, liu2024video,zheng2020image,zheng2021learning,yu2022towards}, greatly outperforming early handcraft feature based approaches~\cite{ sun2014scanned,liu2015moire,yang2017textured,yang2017demoireing}. However, they exhibit poor generalization capability due to insufficient diverse and realistic training data, and researchers have thus begun exploring the potential of synthesized data.

\begin{table*}[ht]
\centering
\small
\setlength{\tabcolsep}{2.37mm}
\scalebox{0.9}{
\begin{tabular}{cc|ccccccc}
\toprule[1.25pt]
\multicolumn{2}{c|}{Datasets} & \multirow{2.5}{*}{Avg. Resolution} & \multirow{2.5}{*}{Size} & \multicolumn{5}{c}{Capture settings}   \\ 
\cmidrule(lr){1-2} \cmidrule(lr){5-9} 
Type & Name &       &      & Phone  & Screen & Multi-zooming rate & Multi-camera / CMOS  & Screen Panel \\ 
\midrule
\midrule
\multirow{4}{*}{\begin{tabular}[c]{@{}c@{}} Moiré Image \\ Dataset\end{tabular}}  
& TIP2018(R) & 256 $\times$ 256   & 135000 & 3     & 3      & \ding{55}(1x-only) & \ding{55}(Main-only) & IPS-only     \\
& FHDMi(R)     & 1024 $\times$ 1024 & 12000  & 3     & 2      & \ding{55}(1x-only) & \ding{55}(Main-only) & IPS-only     \\
& UHDM(R)   & 4328 $\times$ 3248 & 5000   & 3     & 3      & \ding{55}(1x-only) & \ding{55}(Main-only) & IPS-only     \\ 
& LCDMoiré(S) & 1024 $\times$ 1024 & 10200  & -     & -      & -          & -            & -            \\ 
\midrule
\multirow{2.5}{*}{\begin{tabular}[c]{@{}c@{}} Moiré Pattern \\ Dataset\end{tabular}} 
& MoireSpace(R)  & 2160 $\times$ 1286 & 18147  & 3     & 3      & \ding{55}(1x-only) & \ding{55}(Main-only) & IPS-only     \\ 
\cmidrule{2-9}
& \textbf{Ours(R)} & \textbf{3840} $\times$ \textbf{2160} & \textbf{150000} & \textbf{6} & \textbf{6} & \textbf{\checkmark(1x,2x,3x)} & \textbf{\checkmark(Main,Telephoto)} & \textbf{IPS, SVA}  \\ 
\bottomrule[1.25pt]
\end{tabular}
}
\caption{Comparisons of different moiré datasets. The ``R'' denotes the real dataset, and the ``S'' denotes the synthetic dataset.}
\label{tab:datasets-overview}
\end{table*}

\subsection{Moiré Image Synthesis}
An important category focuses on extracting moiré patterns from existing moiré images. Cyclic~\cite{cyclic} and UnDeM~\cite{undem} utilized GAN-based networks to generate moiré images from unpaired real moiré image datasets, resembling moiré patterns found in moiré images while retaining details from moiré-free images. However, they are unstable and constrained by the moiré patterns present in the real image datasets. 
Another category directly simulates moiré patterns on natural images. Shooting~\cite{shooting} simulated the interference of image processing to produce moiré patterns on natural images while Yang et al.~\shortcite{yang2023doing} collected background-independent moiré patterns and then superimposes the natural image with the collected pattern to synthesize moiré images. 
Unfortunately, due to the real-to-synthetic discrepancy, their model performance is limited in real-world applications. 
In contrast, our solution can produce realistic-looking and diverse data to greatly improve demoiréing models' performance.

\subsection{Moiré Dataset} 
TIP18~\cite{sun2018moire}, FHDMi~\cite{he2020fhde}, UHDM~\cite{yu2022towards} are the most widely-used real-world moiré image dataset with increased resolutions 256, 1080P, and 4K, respectively. To lessen the burden of huge human efforts, a synthetic moiré image dataset LCDMoiré~\shortcite{yuan2019aim} has been generated through shooting simulation.
However, synthetic datasets often fail to accurately replicate real imaging processes, making it difficult for demoiréing models trained on them to perform well in real-world situations.
More recently, MoireSpace~\cite{yang2023doing} collects background-independent moiré pattern data for a different task, moiré detection. Inspired by it, we propose to collect a real moiré pattern dataset for image demoiréing. Taking inspiration from this effort, we introduce a real moiré pattern dataset specifically tailored for image demoiréing. Comparatively, our dataset boasts a larger volume and greater diversity of data.

\section{Method}

\subsection{Overview} 
The generalization ability of SOTA demoiréing models is greatly limited by the scarcity of data. Therefore, we mainly face two challenges to obtain a universal model with improved generalization capability: To obtain a vast amount of 1) diverse and 2) realistic-looking moiré data.
Notice that traditional moiré image datasets contain real data, but continuously expanding their size to involve more diversity is extremely time-consuming and impractical. 
While current synthesized datasets/methods struggle to synthesize realistic-looking moiré images.
Hence, to tackle these challenges, we introduce a universal solution, UniDemoiré (Figure~\ref{fig:pipeline}). The data diversity challenge is solved by collecting a more diverse moiré pattern dataset and presenting a moiré pattern generator to increase further pattern variations. 
Meanwhile, the data realistic-looking challenge is undertaken by a moiré image synthesis module.  
Finally, our solution can produce realistic-looking moiré images of sufficient diversity, substantially enhancing the zero-shot and cross-domain performance of demoiréing models.

\subsection{Moiré Pattern Dataset}
\label{subsec: Moiré_Patterns_Collection}

The traditional demoiréing datasets~\cite{sun2018moire,he2020fhde,yu2022towards} typically exhibit a 1-1 correspondence, 1 clean image corresponds to only 1 moiré-contaminated image. However, in the real world, an image may be affected by various moiré patterns.
Meanwhile, aligning moiré images with clean images often introduces errors because of the non-linear distortions and moiré artifacts within cameras.
Therefore, we propose to collect a moiré pattern dataset rather than a moiré image dataset, with no need for image alignment and can easily synthesize multiple moiré counterparts of a single natural image. 
The collection of such a dataset is inspired by MoireSpace, which is designed to address the problem of detecting the presence of moiré rather than to eliminate moiré artifacts.  

\begin{figure}[t]
  \centering
    \includegraphics[width=1.0\linewidth]{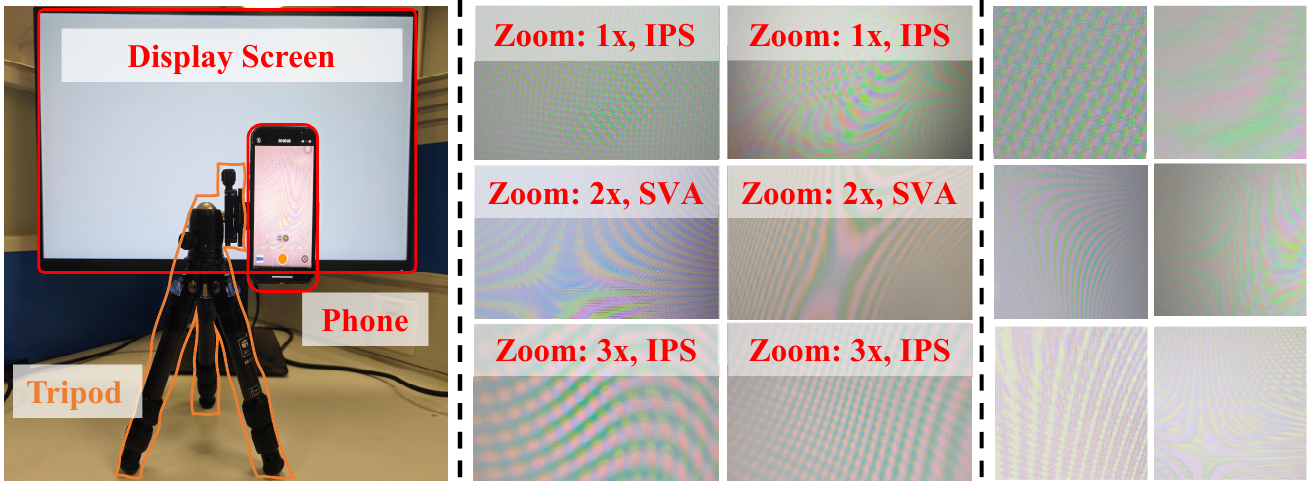}
  \caption{ Data collection setup (left), and examples of moiré patterns in our dataset captured at different zoom rates and screen panel (middle), and our generated patterns (right). } 
  \label{fig:data_collection}
\end{figure}

\paragraph{Capturing Process}
We capture videos of real-world moiré patterns on a pure white screen with a mobile phone to minimize color distortion in the moiré patterns. After recording, frames are uniformly extracted from each video to constitute our dataset. The setup is shown in Figure~\ref{fig:data_collection}-left.

\paragraph{Data Diversity}
To enhance pattern diversity, we build our dataset by considering additional factors that influence moiré formation, which were overlooked in previous moiré datasets, including zooming rate, camera types, CMOS, and screen panel types. Besides, we doubled the number of mobile devices and display screens compared to existing datasets.
A detailed comparison of ours and others is shown in Table~\ref{tab:datasets-overview}.
In summary, our dataset showcases an expanded size, 150000 moiré patterns, in standard 4K resolution with increased diversity. More dataset details are in the appendix. 

\subsection{Moiré Pattern Generation}
\label{subsec: Moiré_Pattern_Generator}
 
Although we have collected a large scale of diverse data, it cannot encompass all conceivable moiré patterns.   
Inspired by recent diffusion models, which have been successfully trained towards diverse image generation in many tasks~\cite{dhariwal2021diffusionmodelsbeatgans}, we propose to use diffusion models to further sample more diverse moiré patterns by sufficiently learning the structural, textural, and color representations of real moiré patterns. 
In this stage, we propose a multi-scale cropping strategy and a colorfulness-sharpness selection strategy to filter high-quality real data. Then we learn the distribution of real moiré patterns in the latent space to generate diverse patterns (Figure~\ref{fig:data_collection}-right).

\begin{figure}[t]
  \centering
    \includegraphics[width=1.0\linewidth]{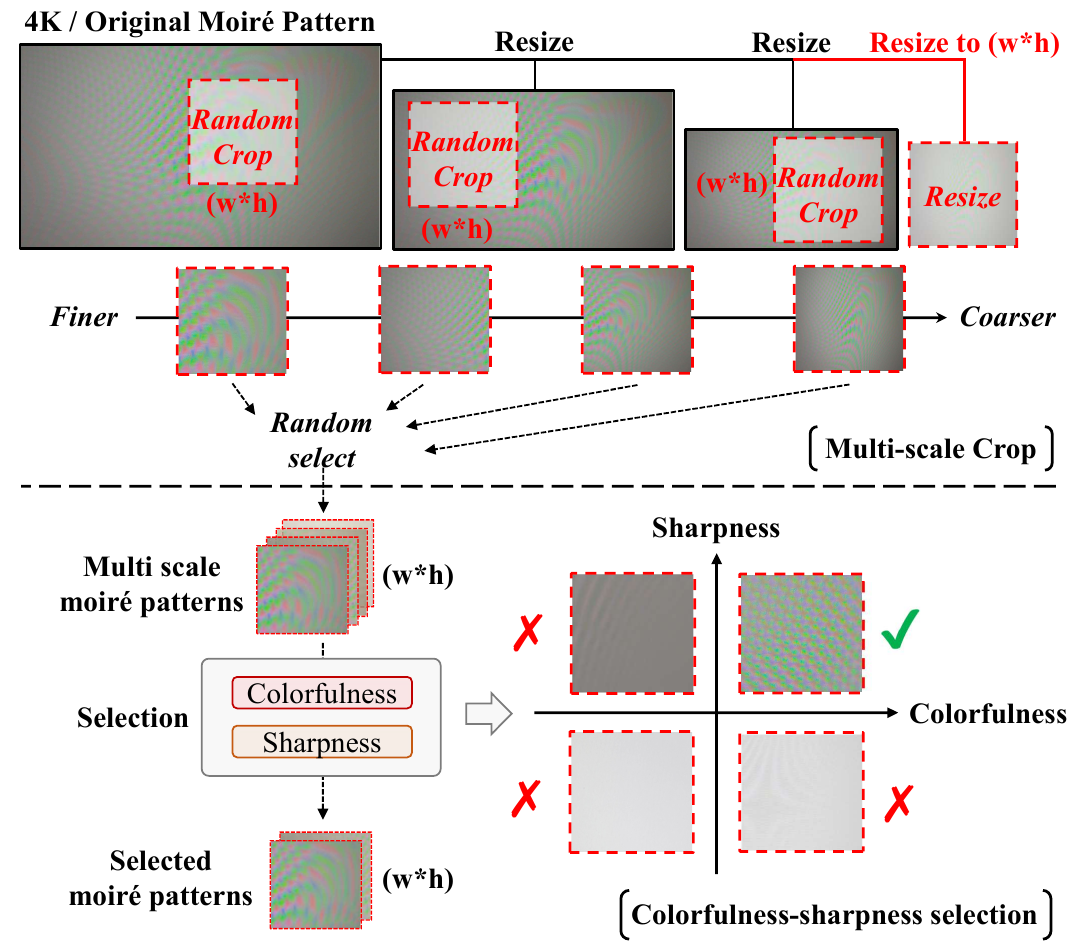}
  \caption{Data preprocessing for moiré pattern generation. }
  \label{fig:generation}
\end{figure}

\paragraph{Multi-Scale Cropping}
Demoiréing models typically employ image patches cropped from the entire image for training. However, given the significant variation of image size in different demoiréing datasets, the scale of content in cropped image patches of the same size also varies greatly. 
Hence, to simulate this process and enhance the diversity of the training data, we perform multi-scale cropping (Figure~\ref{fig:generation} up). 
In particular, 4k images are resized to different sizes, from which we extract and randomly select image patches of uniform size as training data.
In this way, the patches extracted from low and high-resolution images emphasize overall patterns and finer details, respectively.

\paragraph{Colorfulness-Sharpness Selection}
We notice that certain patches involve visually invisible patterns (with a ``\ding{55}'' mark in Figure~\ref{fig:generation}). They potentially confuse the generator during training, aiming to generate moiré pattern images rather than to reproduce plain white images. Hence, we filter out such patches based on colorfulness and sharpness. 
As depicted in Figure~\ref{fig:generation} lower-right, an increased sharpness value indicates more visible moiré patterns, while an increased colorfulness value signifies patterns with richer colors.
The sharpness metric is calculated as the standard deviation of grayscaled input image processed with an edge filter, while the colorfulness metric is calculated as the average standard deviations of A and B channels in image LAB color space.

\subsubsection{Learning Moiré Patterns in the Latent Space} 
As shown in Figure~\ref{fig:data_collection}(middle), plenty of pixels in the moiré pattern appear pure white. 
This leads to a polarization in the pixel distribution of the moiré pattern images, where informative data is concentrated in a few pixels with high values while the rest contains little information.
Based on this observation, we choose to compress the moiré pattern into the latent space through an autoencoder for a more compact and efficient representation of its structural, textural, and color information. For better stability and controllability, we utilize the Latent Diffusion Model~\cite{Rombach2022LDM} to effectively model the complex distribution of the moiré pattern in the latent space.
Examples of generated moiré patterns are shown in Figure~\ref{fig:data_collection} right.
More examples are in the appendix.

\subsection{Moiré Image Synthesis}
\label{subsec: Moiré_Image_Synthesis}

\begin{figure*}[ht]
  \centering
    \includegraphics[width=0.95\linewidth]{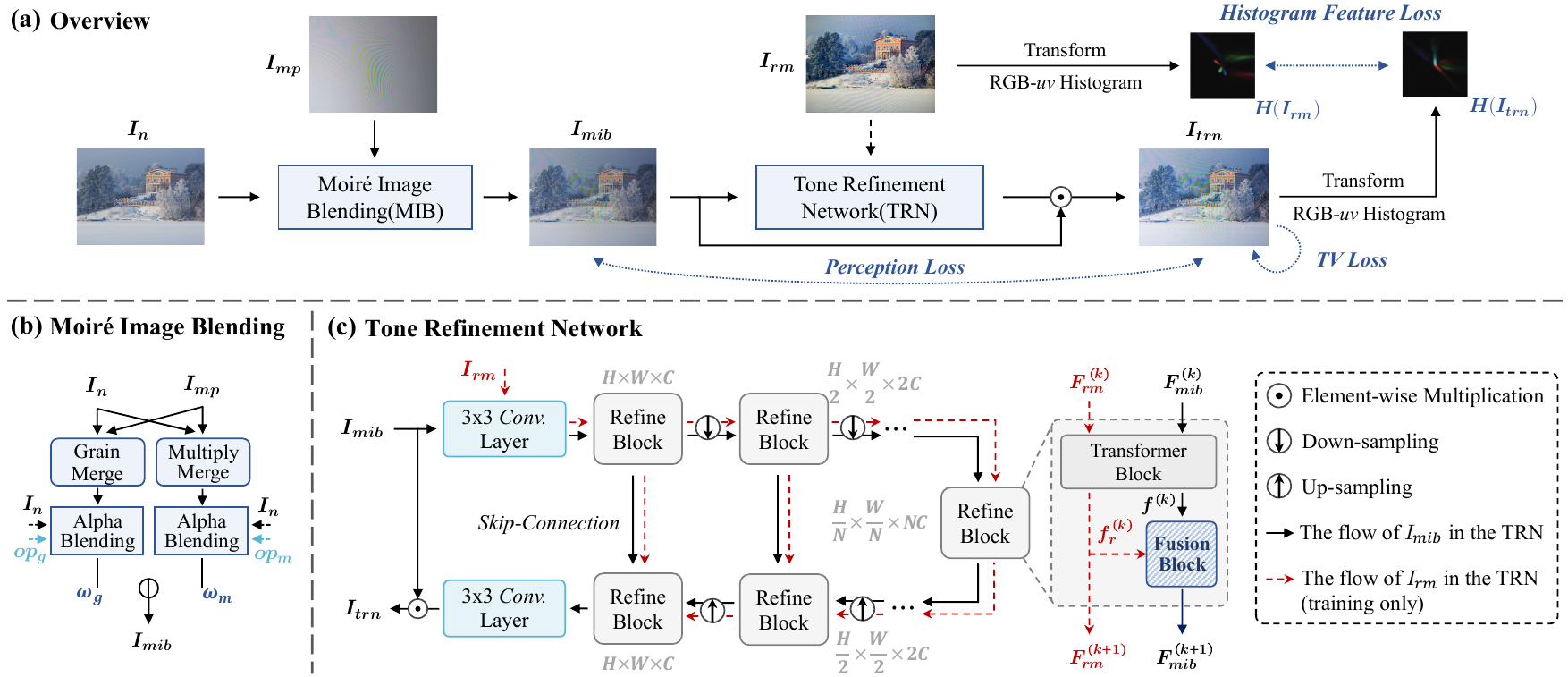}
  \caption{Overview of the Moiré Image Synthesis stage (a). It involves a Moiré Image Blending module (b) for initial moiré image synthesis and a Tone Refinement Network (c) to refine for more realistic results.}
  \label{fig:systhesis}
\end{figure*}

Via data collection and generation, we obtain a vast number of diverse moiré patterns. 
Then, we need to composite moiré patterns with clean images $I_{n}$ to form moiré images. To make the synthesized images realistic-looking,
We first create handcraft rules to produce initial moiré images in the Moiré Image Blending (MIB) module, then design a Tone Refinement Network (TRN) to further faithfully replicate the color and brightness variations observed in real scenes that cannot be fully formulated in those handcraft rules. 
The proposed synthesis process is illustrated in Figure~\ref{fig:systhesis}.

\subsubsection{Moiré Image Blending}
We blend the clean natural image $I_{n}$ (background layer) with the moiré pattern $I_{mp}$ (foreground layer) to form our initial moiré image $I_{mib}$. Notice that MoireSpace~\cite{yang2023doing} synthesized their moiré image $I'_{sm} $via a Multiply Strategy $M(\cdot, \cdot)$,
\begin{equation}
     I'_{sm} = M(I_{mp}, I_{n}) = I_{mp} \odot I_{n},
 \end{equation}
where ``$\odot$'' denotes element-wise multiplication.
However, the result produced by MoireSpace~\cite{yang2023doing} tends to be dark and cannot replicate the desired contrast and color distortion, as shown in Figure~\ref{fig:synthesis_result}. 
Therefore, we design the following handcraft rules to make the blending more realistic (Figure~\ref{fig:systhesis}b). 
we first incorporate an additional blending strategy, Grain Merge ~\cite{LayerModes} $G(\cdot, \cdot)$. Such a brighter strategy can balance the darker result from $M(\cdot, \cdot)$: 
\begin{equation}
    G(I_{mp}, I_{n}) =  I_{mp} + I_{n} - 0.5.
\end{equation} 
Then, we incorporate transparency of the layers using alpha blending~\cite{Porter_Duff_1984} to obtain $I_{comp}^{M}$ and $I_{comp}^{G}$ :
\begin{align}    
I_{comp}^{M} &= r_m \cdot M(I_{mp}, I_{n}) + [1-r_m] \cdot I_{n},\\
I_{comp}^{G} &= r_g \cdot G(I_{mp}, I_{n}) + [1-r_g] \cdot I_{n}.
\end{align}
where $r_m$ and $r_g$ represent the composition ratio parameter of foreground layer $M(I_{mp}, I_{n})$ and $G(I_{mp}, I_{n})$:
\begin{equation}
    r_x = \frac{op_{x}}{op_{x}+(1-op_{x})\cdot op_{n}}, x \in \{m,g\}.
\end{equation}
where $op_m$, $op_g$ represent the opacity of the output layers from the Multiply and the Grain Merge strategies, and $op_{n}$ represent the opacity of the background layer $I_{n}$.
Finally, we perform a weighted (weight $\omega_m$, $\omega_g$) combination of $I_{comp}^{M}$ and $I_{comp}^{G}$ to obtain $I_{mib}$ :
\begin{equation}    
I_{mib} = \omega_m \cdot I_{comp}^{M} + \omega_g \cdot I_{comp}^{G}.
\end{equation}
A visual comparison of MoireSpace result $I'_{sm}$, and our $I_{mib}$ is shown in Figure~\ref{fig:synthesis_result}, showing the superior of $I_{mib}$ over $I'_{sm}$.
Please refer to the appendix for more visual results.

\subsubsection{Tone Refinement Network} 
Though the moiré image blending module creates a preliminary moiré image $I_{mib}$, such a synthesized result based on handcraft rules still struggles to replicate accurate color and brightness changes. 
Comparatively, networks are more powerful in capturing such unknown changes and distortion by progressive learning.
Hence, we present a learnable refinement network to synthesize more realistic results.

The Tone Refinement Network (TRN) proposed here is built on a U-shaped transformer backbone~\cite{Wang2022Uformer} incorporating multiple refine blocks, illustrated in Figure~\ref{fig:systhesis} (c). 
It takes $I_{mib}$ as input, applies pixel-wise tone adjustment to $I_{mib}$, and minimizes the tone gap between the output $I_{trn}$ and the given real moiré images $I_{rm}$. 
To be clear,
TRN firstly applies a 3$\times$3 convolutional layer with LeakyReLU to extract tone features $F_{mib}^{(0)}$, $F_{rm}^{(0)}$. 
Next, the feature maps $F_{mib}^{(0)}$ and $F_{rm}^{(0)}$ are passed through $N$ encoder phases and $N$ decoder phases with skip connections. 
Each phase contains a refine block to capture long-range dependencies, benefiting from the self-attention in Transformer. 

\begin{figure}[t]
  \centering
    \includegraphics[width=1.0\linewidth]{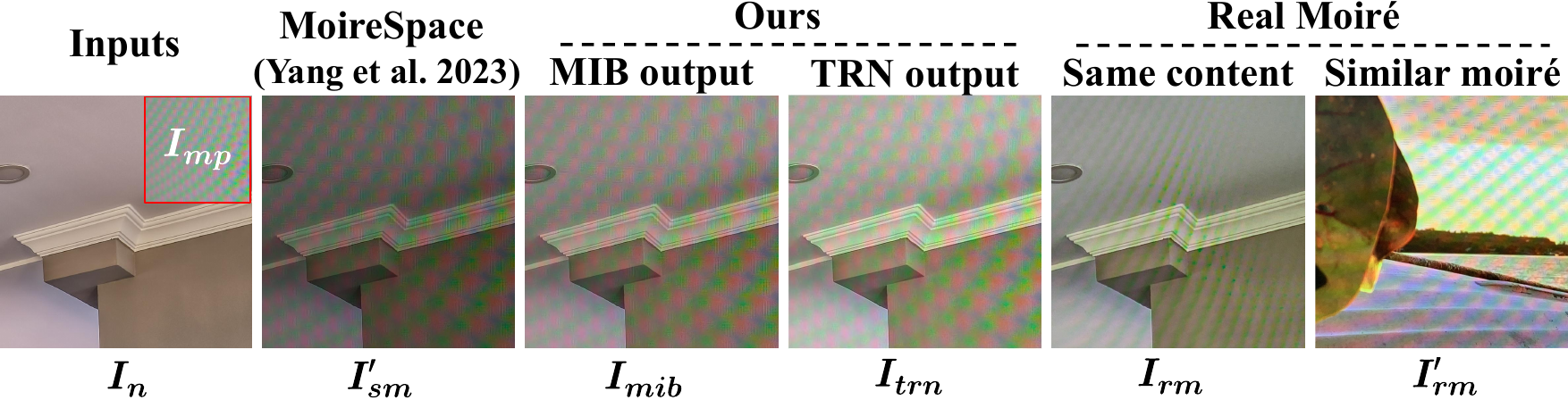}
  \caption{Visualization of our intermediate synthetic results.  
  } 
  \label{fig:synthesis_result}
\end{figure}

Inspired by research in style transfer and domain generalization~\cite{Ulyanov2016IN, Huang2017Arbitrary, Zhou2021Mixstyle}, we design a tone feature fusion block within each refine block to better fuse the tone feature statistics between $I_{mib}$ and corresponding $I_{rm}$. 
It mixes the feature statistics of two instances with a random convex weight. 
As illustrated in Figure~\ref{fig:systhesis} (c), the computations inside a fusion block module in the $k$-th refine block can be summarized into two steps. 
First, given two sets of feature maps $f^{(k)}$ and $f_r^{(k)}$ for $I_{mib}$ and $I_{rm}$, the fusion block generates a mixture of feature statistics, 
\begin{align}
    \gamma_{mix} &= \lambda \cdot \sigma(f^{(k)}) + (1-\lambda) \cdot \sigma(f^{(k)}_r), \\
    \beta_{mix}  &= \lambda \cdot \mu(f^{(k)})    + (1-\lambda) \cdot \mu(f^{(k)}_r).
\end{align}
where $\mu$ and $\sigma$ represent the mean and variance of feature maps, while $\lambda$ is a random weight sampled from the beta distribution, $\lambda \in \text{Beta}(\alpha, \alpha)$ with $\alpha \in (0, \infty)$ being a hyper-parameter.
Then, the mixture of feature statistics is applied to the tone-normalized $F_{mib}^{(k+1)}$:
\begin{equation}
    F_{mib}^{(k+1)} = \gamma_{mix} \odot \frac{f^{(k)}-\mu(f^{(k)})}{\sigma(f^{(k)})} + \beta_{mix}.
\end{equation}
The fusion block can effectively utilize the moiré feature information of $I_{rm}$ and greatly helps reduce the moiré domain gap between the final synthesized image $I_{trn}$ and real moiré image $I_{rm}$, which is one significant innovation.
After the $N$ decoder stages, we apply a 3$\times$3 convolution layer on feature maps $F_{mib}^{(2N)}$ to obtain a tone refinement matrix $M_{trn}$. 
Finally, the synthetic image is obtained by $I_{trn}=I_{mib} \odot M_{trn}$ after color normalization, where ``$\odot$'' represents element-wise multiplication. 
Notice that the fusion block is solely utilized in the training phase, and $I_{rm}$ is exclusively fed into the network during training. 
Figure~\ref{fig:synthesis_result} compares the initial blending result $I_{mib}$ with the final synthesized result $I_{trn}$. Please refer to the appendix for more results.

\begin{table*}[ht]
\small
\centering
\setlength{\tabcolsep}{2.26mm}
\scalebox{0.9}{
\begin{tabular}{cc|ccccc|ccccc}
\toprule[1.25pt]
\multirow{2.5}{*}{\makecell{Test \\ Dataset}} & \multirow{2.5}{*}{Metric} & \multicolumn{5}{c|}{Demoiréing Network: \textbf{MBCNN}}  & \multicolumn{5}{c}{Demoiréing Network:   \textbf{ESDNet-L}}  \\ 
 \cmidrule(lr){3-7}  \cmidrule(lr){8-12}
&           & Shooting & UnDeM$^\dagger$ & UnDeM$^\ddagger$ & MoireSpace  & Ours & Shooting & UnDeM$^\dagger$ & UnDeM$^\ddagger$ & MoireSpace & Ours \\ 
\midrule
\midrule
 \multirow{3}{*}{UHDM}  &\ua{PSNR } & 9.2284  & 13.4256 & 14.5237 & 14.7826 & \textbf{17.9162} & 10.2568 & 15.2269 & 15.2947 & 14.7989 & \textbf{17.2524} \\
                        &\ua{SSIM } & 0.5180  & 0.3973  & 0.4425  & 0.4724  & \textbf{0.6280}  & 0.5664  & 0.5873  & 0.5777  & 0.4859  & \textbf{0.6454}  \\
                        &\da{LPIPS }& 0.6664  & 0.6489  & 0.6332  & 0.5568  & \textbf{0.4162}  & 0.5130  & 0.4190  & 0.4241  & 0.5254  & \textbf{0.3238}  \\ 
\cmidrule{1-12} 
 \multirow{3}{*}{FHDMi} &\ua{PSNR } & 10.6750  & 17.8355 & 18.1652 & 18.5523 & \textbf{19.0094} & 11.6022 & 18.4335 & 18.5390 & 18.0763 & \textbf{19.8128} \\
                        &\ua{SSIM } & 0.4478  & 0.6802  & 0.6999  & 0.7094  & \textbf{0.7137}  & 0.5425  & 0.6900  & 0.6812  & 0.7189  & \textbf{0.7319}  \\
                        &\da{LPIPS }& 0.5978  & 0.2606  & 0.2472  & 0.2742  & \textbf{0.2390}  & 0.4515  & 0.2877  & 0.2986  & 0.2616  & \textbf{0.2134}  \\ 
\bottomrule[1.25pt]
\end{tabular}
}
\caption{Quantitative results of zero-shot demoiréing trained with synthesized data only. 
``$\dagger$'' indicates UnDem uses moiré patterns retrieved from real data in TIP for inference.  ``$\ddagger$''  indicates UnDem uses our generated moiré pattern for inference.
}
\label{tab:Exp_multi_datasets}
\end{table*}

\begin{table*}[t]
\small
\centering
\setlength{\tabcolsep}{1.89mm}
\scalebox{0.9}{
\begin{tabular}{ccc|ccccc|ccccc}
\toprule[1.25pt]    

\multicolumn{2}{c}{Cross Dataset}& \multirow{2.5}{*}{Metric} & \multicolumn{5}{c|}{Demoiréing Network: \textbf{MBCNN}}  & \multicolumn{5}{c}{Demoiréing Network:   \textbf{ESDNet-L}}  \\ 
\cmidrule(lr){1-2} \cmidrule(lr){4-8}  \cmidrule(lr){9-13}

Source & Target &        & Baseline & Shooting & UnDeM & MoireSpace & Ours  & Baseline & Shooting & UnDeM & MoireSpace   & Ours   \\ 
\midrule
\midrule
\multirow{6.5}{*}{UHDM}    
& \multirow{3}{*}{FHDMi} &\ua{PSNR } & 19.3848 & 19.2032 & 19.4676 & 19.4531 & \textbf{19.8625} & 20.3422 & 20.2407 & 20.4014 & 20.2806 & \textbf{20.7543} \\
&                        &\ua{SSIM } & 0.7436  & 0.7459  & 0.7455  & 0.7496  & \textbf{0.7525}  & 0.7599  & 0.7579  & 0.7510  & 0.7603  & \textbf{0.7653}  \\
&                        &\da{LPIPS }& 0.3002  & 0.2975  & 0.2964  & 0.2993  & \textbf{0.2842}  & 0.2525  & 0.2632  & 0.2509  & 0.2324  & \textbf{0.2136}  \\ 
\cmidrule{2-13} 
& \multirow{3}{*}{TIP}   &\ua{PSNR } & 17.8107 & 18.3730 & 18.6674 & 18.9214 & \textbf{19.3922} & 18.8040 & 18.4543 & 19.3545 & 19.3964 & \textbf{19.5009} \\
&                        &\ua{SSIM } & 0.6627  & 0.6888  & 0.6911  & 0.6996  & \textbf{0.7022}  & 0.6921  & 0.6930  & 0.6998  & 0.7111  & \textbf{0.7149}  \\
&                        &\da{LPIPS }& 0.3580  & 0.3886  & 0.3909  & 0.3829  & \textbf{0.3781}  & 0.3524  & 0.3849  & 0.3601  & 0.3522  & \textbf{0.3495}  \\ 
\midrule
\multirow{6.5}{*}{FHDMi}   
& \multirow{3}{*}{UHDM}  &\ua{PSNR } & 17.1331 & 17.5326 & 17.4870 & 17.6050 & \textbf{18.7931} & 18.0049 & 18.4189 & 17.9574 & 17.9751 & \textbf{18.9240} \\
&                        &\ua{SSIM } & 0.6159  & 0.6334  & 0.6331  & 0.6642  & \textbf{0.7186}  & 0.5755  & 0.5780  & 0.5857  & 0.5548  & \textbf{0.6658}  \\
&                        &\da{LPIPS }& 0.4470  & 0.4350  & 0.4285  & 0.4020  & \textbf{0.3508}  & 0.4420  & 0.4279  & 0.4460  & 0.4579  & \textbf{0.3405}  \\ 
\cmidrule{2-13} 
& \multirow{3}{*}{TIP}   &\ua{PSNR } & 20.2161 & 20.7793 & 20.8261 & 20.1194 & \textbf{21.0694} & 20.6647 & 20.8678 & 20.4663 & 20.8107 & \textbf{21.5786} \\
&                        &\ua{SSIM } & 0.7340  & 0.7304  & 0.7381  & 0.7347  & \textbf{0.7494}  & 0.7504  & 0.7606  & 0.7278  & 0.7582  & \textbf{0.7668}  \\
&                        &\da{LPIPS }& 0.2979  & 0.2884  & 0.2891  & 0.2961  & \textbf{0.2832}  & 0.2459  & 0.2450  & 0.2998  & 0.2468  & \textbf{0.2310}  \\ 
\midrule
\multirow{6.5}{*}{TIP}     
& \multirow{3}{*}{UHDM}  &\ua{PSNR } & 17.3409 & 17.4011 & 17.4407 & 17.4987 & \textbf{18.2937} & 17.4332 & 16.1836 & 16.8402 & 16.6296 & \textbf{18.4978} \\
&                        &\ua{SSIM } & 0.6144  & 0.6062  & 0.6066  & 0.6059  & \textbf{0.6913}  & 0.5523  & 0.5511  & 0.5692  & 0.5748  & \textbf{0.6866}  \\
&                        &\da{LPIPS }& 0.4726  & 0.4487  & 0.4473  & 0.4412  & \textbf{0.3990}  & 0.4987  & 0.4723  & 0.4532  & 0.4387  & \textbf{0.3231}  \\ 
\cmidrule{2-13} 
& \multirow{3}{*}{FHDMi} &\ua{PSNR } & 18.9458 & 19.2731 & 19.0336 & 19.1101 & \textbf{20.1053} & 19.2368 & 18.1936 & 19.2112 & 18.8385 & \textbf{19.9971} \\
&                        &\ua{SSIM } & 0.7369  & 0.7399  & 0.7215  & 0.7321  & \textbf{0.7725}  & 0.7354  & 0.7297  & 0.7499  & 0.7389  & \textbf{0.7580}  \\
&                        &\da{LPIPS }& 0.2494  & 0.2447  & 0.2452  & 0.2382  & \textbf{0.2315}  & 0.2316  & 0.2320  & 0.2130  & 0.2228  & \textbf{0.1915}  \\ 
\bottomrule[1.25pt]  
\end{tabular}
}
\caption{Quantitative results of cross-dataset evaluations.}
\label{tab:Exp_cross_datasets}
\end{table*}

\subsubsection{Loss Functions}
The tone adjustment network aims to adjust the overall color tone and contrast of $I_{trn}$ in a way that it resembles $I_{rm}$ without affecting moiré pattern $I_{mp}$. 

First, moiré patterns can disrupt image structures by generating strip-shaped artifacts~\cite{yu2022towards}. Therefore, comparing two moiré images directly in pixel space is less effective. 
Thus, we adopt the perceptual loss~\cite{johnson2016perceptual} $\mathcal{L}_{per}$ to 
optimize the $\mathcal{L}_1$ distance between the extracted content features of $I_{mib}$ and $I_{trn}$:
\begin{small}
\begin{equation}
\mathcal{L}_{per}(I_{trn}, I_{mib})=\sum_{j=1}^{N_L}\frac{\left\|\phi_j(I_{trn})-\phi_j(I_{mib})\right\|_1}{C_j H_j W_j}, 
\end{equation}
\end{small}
\normalsize 
where $\phi_j(I)$ is the activations of the $j$-th layer of the VGG16 network~\cite{simonyan2014very}, and $N_L$ represents the number of convolutional layers in VGG16.

In addition, to effectively tune the tone of $I_{trn}$, we take advantage of color differentiable RGB-uv histogram features $H(I_{trn})$ and $H(I_{rm})$ in log chromaticity space, inspired by color constancy method~\cite{barron2015ccc, Afifi2021histogan}, 
as shown in Figure~\ref{fig:systhesis} (a). Such RGB-uv histograms have proven efficient in color transfer tasks~\cite{Eibenberger2012log}. 
We optimize color loss 
using the differentiable Hellinger distances 
\begin{equation}
    \mathcal{L}_{color}\left(I_{trn}, I_{rm}\right)=\left\|H(I_{trn})^{1 / 2}-H(I_{rm})^{1 / 2}\right\|_2,
\end{equation}
where $\left\| \cdot \right\|_2$ is the standard Euclidean norm and $\cdot ^{1/2}$ is an element-wise square root. 

Finally, 
we use total variation regularizer $\mathcal{L}_{tv}$ to remove unwanted details while encouraging spatial smoothness: 
\begin{small}
\begin{equation}
\mathcal{L}_{tv}(I_{trn})=\sum_{i=1}^H \sum_{j=1}^W\left|I_{trn}^{i+1, j}-I_{trn}^{i, j}\right|+\left|I_{trn}^{i, j+1}-I_{trn}^{i, j}\right|
\end{equation}
\end{small}
\normalsize

Total loss $\mathcal{L}$ is then defined as a weighted compound of $\mathcal{L}_{per}$, $\mathcal{L}_{color}$ and $\mathcal{L}_{tv}$:
\begin{equation}
    \mathcal{L} = \lambda_{per}\mathcal{L}_{per} + \lambda_{color}\mathcal{L}_{color} + \lambda_{tv}\mathcal{L}_{tv}.
\end{equation}

\subsection{Image Demoiréing}
\label{subsec: Image_Dmoiréing}
Our contributions mainly lie in the above three stages. 
Then, diverse and realistic-looking data synthesized by our solution can be seamlessly integrated with demoiréing models to improve their performance.

\section{Experiments}
\subsection{Experimental Setups} 
For all compared methods, we used their released code. Thorough implementation details are in the appendix. 

\begin{figure}[t]
  \centering
\includegraphics[width=1.0\linewidth]{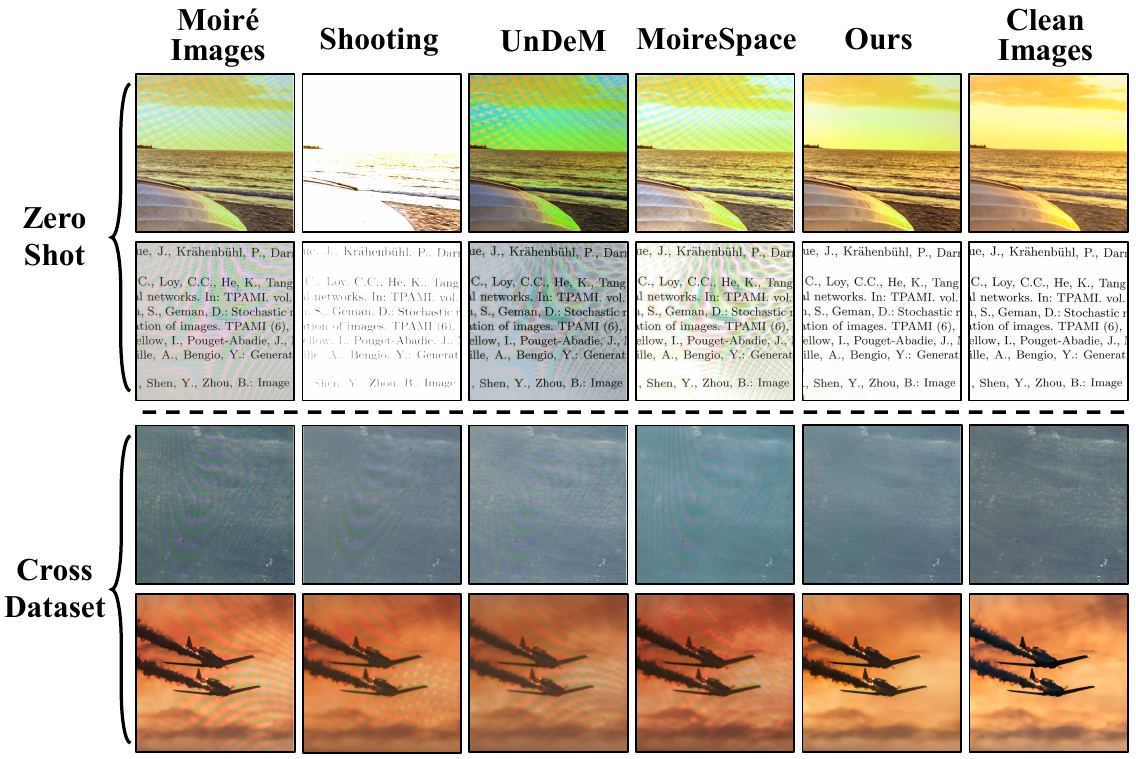}
  \caption{Comparisons of demoiréing results.} 
  \label{fig:demoiré_result}
\end{figure}

\subsubsection{Datasets and Metrics.}
\textbf{\emph{1) }Moiré Pattern Dataset} is used to train our moiré pattern generator. 
\textbf{\emph{2) }Real Moiré Image Dataset}, TIP~\cite{sun2018moire}, FHDMi~\cite{he2020fhde}, and UHDM~\cite{yu2022towards}, are used to demonstrate our ability in restoring real moiré images. 
\textbf{\emph{3) }Evaluation Metrics.}  We evaluate demoiréing performance on the Peak-Signal-to-Noise Ratio (PSNR), Structural Similarity Index (SSIM)~\cite{wang2004image}, and LPIPS~\cite{zhang2018unreasonable}.

\subsubsection{Comparison Methods}
We compare UniDemoiré to the SOTA synthesis methods in 3 current modalities: the simulation method ``Shooting"~\cite{shooting}, the implicit moiré synthesis approach ``UnDeM"~\cite{undem}, which employs a neural network, and the explicit synthesis method termed ``MoireSpace"~\cite{yang2023doing}, which utilizes its moiré pattern dataset.

\subsubsection{Demoiréing Models}
We test on the most effective SOTA demoiréing models,  MBCNN~\cite{zheng2020image} and ESDNet-L~\cite{yu2022towards}.

\subsection{Zero-Shot Demoiréing with Synthesized Data Only} 
We first demonstrate demoiréing results on real moiré images trained on purely synthesized data by SOTA moiré synthesis methods. To avoid data overlap in training sets and test sets, we have collected a comprehensive Mixed High-Resolution Natural Image Dataset (MHRNID), based on which, moiré images are synthesized for training demoiréing models. 
Quantitative comparisons can be found in Table~\ref{tab:Exp_multi_datasets}. Visual comparisons on demoiréing real data in UHDM are illustrated in Figure~\ref{fig:demoiré_result}.
Due to that UnDeM relies on existing moiré images in both the training (fusion networks) and inference phase, 
we trained their networks on the TIP dataset and showed the result of UnDeM using the real moiré in the TIP dataset (``$\dagger$" in Table  \ref{tab:Exp_multi_datasets}) and our sampled moiré pattern (``$\ddagger$" in Table \ref{tab:Exp_multi_datasets}) during inference, respectively.
For a fair comparison, we also use real moiré images from TIP dataset to train our TRN.
Notice that UnDeM and our method only use real moiré images to guide the synthesis, and neither of us uses such real data to train demoiréing models directly.

From the quantitative perspective (Table~\ref{tab:Exp_multi_datasets}), our method substantially outperforms all other approaches, particularly by more than 3.2 dB and 2.0 dB for MBCNN and ESDNet-L on the UHDM dataset,  respectively.
Besides, UnDeM$^\ddagger$ using our generated moiré patterns outperforms UnDeM$^\dagger$ using real moiré patterns in all experiments, proving our effectiveness further. 
From the qualitative perspective (Figure~\ref{fig:demoiré_result}),  our method demonstrates strong capability even when images in the target domain are contaminated by severe moiré patterns, which other synthesis methods fail to address.
We attribute our superiority to the diversity and realism of our synthetic data. Such high-quality data by our UniDemoiré enables the demoiréing model to learn moiré characteristics better, improving performance in removing unseen moiré artifacts. 
More visual results are in the appendix.

\subsection{Cross-Dataset Evaluation} 
We then demonstrate our ability to improve the performance of demoiréing models across domains. 
Quantitative results are shown in Table ~\ref{tab:Exp_cross_datasets}. Note that ``Baseline" means that the demoiréing models (MBCNN and ESDNet-L) are trained with the original source real moiré datasets and tested on the target dataset. 
For each synthesis approach, a demoiréing model is trained with combined original real data in the source dataset and corresponding synthesized data. 

As shown, the Shooting method struggles with real data due to differences between synthetic and real moiré. 
UnDeM relies on a GAN network but can be inconsistent depending on the dataset and quality. 
The MoireSpace method performs better than UnDeM but has inferior moiré patterns and synthesis quality, resulting in lower experimental metrics. 
Thanks to the realistic and diverse synthesized data, our method outperforms all previous methods across every experiment.
Visual comparisons in Figure~\ref{fig:demoiré_result} (lower, Source: UHDM, Target: FHDMi) demonstrate our effectiveness. 

\begin{table}[t]
\centering
\small
\setlength{\tabcolsep}{2mm}
\begin{tabular}{lccc}
\toprule
Components                     &\ua{PSNR}   &\ua{SSIM} &\da{LPIPS} \\ 
\midrule
ALL                          & \textbf{20.7543} & \textbf{0.7653} & \textbf{0.2136} \\

$w/o$ MPG                    & 20.1607          & 0.7326          & 0.2456          \\
$w/o$ TRN                    & 20.1691          & 0.7372          & 0.2544          \\
TRN ($w/o$ $\mathcal{L}_{per}$)    & 20.3076          & 0.7508          & 0.2558          \\
TRN ($w/o$ $\mathcal{L}_{color}$)  & 20.2692          & 0.7406          & 0.2301          \\
TRN ($w/o$ $\mathcal{L}_{tv}$)     & 20.3961          & 0.7451          & 0.2324          \\
TRN ($w/o$ fusion block)           & 20.2868          & 0.7370          & 0.2311          \\ 
\bottomrule
\end{tabular}
\caption{Ablation studies. Source: UHDM, Target: FHDMi.}
\label{tab:Exp_ablation}
\end{table}

\subsection{Ablation Study}
We individually ablate submodules in our proposed method to analyze their contribution. All these experiments are trained with the UHDM dataset and validated on the FHDMi dataset. Experimental results in Table~\ref{tab:Exp_ablation} verify that all components in our UniDemoiré solution are crucial for achieving the desired demoiréing performance. Removing any component such as the Moiré Pattern Generator (MPG), Tone Refinement Network (TRN), loss functions, and feature fusion block leads to a significant performance decline. 
More ablation studies are provided in the appendix.

\section{Conclusion}
By addressing the issue of data diversity and realism, our universal solution, UniDemoiré, tackles one of the most important bottlenecks in image demoiréing problems. It showcases significant performance in zero-shot demoiréing and demonstrates a strong capability of enhancing the cross-domain performance of existing demoiréing models. 
More importantly, our method holds the potential to generate billions of moiré data and to significantly expand demoiréing models with a vast increase in parameters. 
Our limitations are discussed in the appendix.

\section{Acknowledgments}
This work is supported by NSFC (No.62206173), Shanghai Frontiers Science Center of Human-centered Artificial Intelligence (ShangHAI), MoE Key Laboratory of Intelligent Perception and Human-Machine Collaboration (KLIP-HuMaCo).
This work is also partially supported by HKU-SCF FinTech Academy,
HKRGC Theme-based research scheme project T35-710/20-R, and SZ-HK-Macau Technology Research Programme \#SGDX20210823103537030.

\bibliography{aaai25}

\begin{thebibliography}{59}
\providecommand{\natexlab}[1]{#1}

\bibitem[{Afifi and Brown(2019)}]{afifi2019sensor}
Afifi, M.; and Brown, M.~S. 2019.
\newblock Sensor-Independent Illumination Estimation for DNN Models.
\newblock arXiv:1912.06888.

\bibitem[{Afifi, Brubaker, and Brown(2021)}]{Afifi2021histogan}
Afifi, M.; Brubaker, M.~A.; and Brown, M.~S. 2021.
\newblock HistoGAN: Controlling Colors of GAN-Generated and Real Images via Color Histograms.
\newblock In \emph{2021 IEEE/CVF Conference on Computer Vision and Pattern Recognition (CVPR)}.

\bibitem[{Afifi et~al.(2019)Afifi, Price, Cohen, and Brown}]{Afifi2019CVPR}
Afifi, M.; Price, B.; Cohen, S.; and Brown, M.~S. 2019.
\newblock When Color Constancy Goes Wrong: Correcting Improperly White-Balanced Images.
\newblock In \emph{Proceedings of the IEEE/CVF Conference on Computer Vision and Pattern Recognition (CVPR)}.

\bibitem[{Amidror(2009)}]{amidror2009theory}
Amidror, I. 2009.
\newblock \emph{The Theory of the Moir{\'e} Phenomenon: Volume I: Periodic Layers}, volume~38.
\newblock Springer Science \& Business Media.

\bibitem[{Barron(2015)}]{barron2015ccc}
Barron, J.~T. 2015.
\newblock Convolutional Color Constancy.
\newblock arXiv:1507.00410.

\bibitem[{Dhariwal and Nichol(2021)}]{dhariwal2021diffusionmodelsbeatgans}
Dhariwal, P.; and Nichol, A. 2021.
\newblock Diffusion Models Beat GANs on Image Synthesis.
\newblock arXiv:2105.05233.

\bibitem[{Dosovitskiy and Brox(2016)}]{Dosovitskiy2016Generating}
Dosovitskiy, A.; and Brox, T. 2016.
\newblock Generating Images with Perceptual Similarity Metrics based on Deep Networks.
\newblock \emph{Neural Information Processing Systems,Neural Information Processing Systems}.

\bibitem[{Eibenberger and Angelopoulou(2012)}]{Eibenberger2012log}
Eibenberger, E.; and Angelopoulou, E. 2012.
\newblock The importance of the normalizing channel in log-chromaticity space.
\newblock In \emph{2012 19th IEEE International Conference on Image Processing}.

\bibitem[{Esser, Rombach, and Ommer(2021)}]{Esser2021Taming}
Esser, P.; Rombach, R.; and Ommer, B. 2021.
\newblock Taming Transformers for High-Resolution Image Synthesis.
\newblock In \emph{2021 IEEE/CVF Conference on Computer Vision and Pattern Recognition (CVPR)}.

\bibitem[{Fei et~al.(2023)Fei, Lyu, Pan, Zhang, Yang, Luo, Zhang, and Dai}]{fei2023generative}
Fei, B.; Lyu, Z.; Pan, L.; Zhang, J.; Yang, W.; Luo, T.; Zhang, B.; and Dai, B. 2023.
\newblock Generative Diffusion Prior for Unified Image Restoration and Enhancement.
\newblock In \emph{Proceedings of the IEEE/CVF Conference on Computer Vision and Pattern Recognition}, 9935--9946.

\bibitem[{GIMP(2023)}]{LayerModes}
GIMP. 2023.
\newblock Layer Blending Modes.
\newblock [Online; accessed 10-April-2024].

\bibitem[{He et~al.(2019)He, Wang, Shi, and Duan}]{he2019mop}
He, B.; Wang, C.; Shi, B.; and Duan, L.-Y. 2019.
\newblock Mop moire patterns using mopnet.
\newblock In \emph{Proceedings of the IEEE/CVF International Conference on Computer Vision}, 2424--2432.

\bibitem[{He et~al.(2020)He, Wang, Shi, and Duan}]{he2020fhde}
He, B.; Wang, C.; Shi, B.; and Duan, L.-Y. 2020.
\newblock FHDe 2 Net: Full High Definition Demoireing Network.
\newblock In \emph{Computer Vision--ECCV 2020: 16th European Conference, Glasgow, UK, August 23--28, 2020, Proceedings, Part XXII 16}, 713--729. Springer.

\bibitem[{Hosu et~al.(2024)Hosu, Agnolucci, Wiedemann, and Iso}]{hosu2024uhdiqa}
Hosu, V.; Agnolucci, L.; Wiedemann, O.; and Iso, D. 2024.
\newblock UHD-IQA Benchmark Database: Pushing the Boundaries of Blind Photo Quality Assessment.
\newblock arXiv:2406.17472.

\bibitem[{Huang and Belongie(2017)}]{Huang2017Arbitrary}
Huang, X.; and Belongie, S. 2017.
\newblock Arbitrary Style Transfer in Real-time with Adaptive Instance Normalization.
\newblock In \emph{2017 IEEE International Conference on Computer Vision (ICCV)}.

\bibitem[{Johnson, Alahi, and Fei-Fei(2016)}]{johnson2016perceptual}
Johnson, J.; Alahi, A.; and Fei-Fei, L. 2016.
\newblock Perceptual losses for real-time style transfer and super-resolution.
\newblock In \emph{European conference on computer vision}, 694--711. Springer.

\bibitem[{Kingma and Ba(2014)}]{kingma2014adam}
Kingma, D.~P.; and Ba, J. 2014.
\newblock Adam: A method for stochastic optimization.
\newblock \emph{arXiv preprint arXiv:1412.6980}.

\bibitem[{Lee et~al.(2021)Lee, Son, Rim, Cho, and Lee}]{lee2021iterative}
Lee, J.; Son, H.; Rim, J.; Cho, S.; and Lee, S. 2021.
\newblock Iterative Filter Adaptive Network for Single Image Defocus Deblurring.
\newblock In \emph{Proceedings of the IEEE/CVF Conference on Computer Vision and Pattern Recognition}, 2034--2042.

\bibitem[{Li et~al.(2021)Li, Li, Zhao, and Xu}]{li2021dehazeflow}
Li, H.; Li, J.; Zhao, D.; and Xu, L. 2021.
\newblock Dehazeflow: Multi-scale conditional flow network for single image dehazing.
\newblock In \emph{Proceedings of the 29th ACM International Conference on Multimedia}, 2577--2585.

\bibitem[{Liu, Yang, and Yue(2015)}]{liu2015moire}
Liu, F.; Yang, J.; and Yue, H. 2015.
\newblock Moir{\'e} pattern removal from texture images via low-rank and sparse matrix decomposition.
\newblock In \emph{2015 Visual Communications and Image Processing (VCIP)}, 1--4. IEEE.

\bibitem[{Liu et~al.(2024)Liu, An, Yuan, Zhou, Li, Wang, and Tian}]{liu2024video}
Liu, L.; An, J.; Yuan, S.; Zhou, W.; Li, H.; Wang, Y.; and Tian, Q. 2024.
\newblock Video Demoir{\'e}ing with Deep Temporal Color Embedding and Video-Image Invertible Consistency.
\newblock \emph{IEEE Transactions on Multimedia}.

\bibitem[{Liu et~al.(2020)Liu, Liu, Yuan, Slabaugh, Leonardis, Zhou, and Tian}]{liu2020wavelet}
Liu, L.; Liu, J.; Yuan, S.; Slabaugh, G.; Leonardis, A.; Zhou, W.; and Tian, Q. 2020.
\newblock Wavelet-based dual-branch network for image demoir{\'e}ing.
\newblock In \emph{Computer Vision--ECCV 2020: 16th European Conference, Glasgow, UK, August 23--28, 2020, Proceedings, Part XIII 16}, 86--102. Springer.

\bibitem[{Loshchilov and Hutter(2016)}]{loshchilov2016sgdr}
Loshchilov, I.; and Hutter, F. 2016.
\newblock Sgdr: Stochastic gradient descent with warm restarts.
\newblock \emph{arXiv preprint arXiv:1608.03983}.

\bibitem[{Loshchilov and Hutter(2019)}]{loshchilov2019adamw}
Loshchilov, I.; and Hutter, F. 2019.
\newblock Decoupled Weight Decay Regularization.
\newblock arXiv:1711.05101.

\bibitem[{Luo et~al.(2020)Luo, Zhang, Hong, Qu, Xie, and Li}]{luo2020deep}
Luo, X.; Zhang, J.; Hong, M.; Qu, Y.; Xie, Y.; and Li, C. 2020.
\newblock Deep wavelet network with domain adaptation for single image demoireing.
\newblock In \emph{Proceedings of the IEEE/CVF Conference on Computer Vision and Pattern Recognition Workshops}, 420--421.

\bibitem[{Luo et~al.(2023)Luo, Gustafsson, Zhao, Sj{\"o}lund, and Sch{\"o}n}]{luo2023refusion}
Luo, Z.; Gustafsson, F.~K.; Zhao, Z.; Sj{\"o}lund, J.; and Sch{\"o}n, T.~B. 2023.
\newblock Refusion: Enabling large-size realistic image restoration with latent-space diffusion models.
\newblock In \emph{Proceedings of the IEEE/CVF Conference on Computer Vision and Pattern Recognition}, 1680--1691.

\bibitem[{Niu, Guo, and Wang(2021)}]{shooting}
Niu, D.; Guo, R.; and Wang, Y. 2021.
\newblock Mori{\'e} attack (ma): A new potential risk of screen photos.
\newblock \emph{Advances in Neural Information Processing Systems}, 34: 26117--26129.

\bibitem[{Niu et~al.(2023)Niu, Lin, Liu, and Guo}]{niu2023progressive}
Niu, Y.; Lin, Z.; Liu, W.; and Guo, W. 2023.
\newblock Progressive Moire Removal and Texture Complementation for Image Demoireing.
\newblock \emph{IEEE Transactions on Circuits and Systems for Video Technology}.

\bibitem[{Odena, Dumoulin, and Olah(2016)}]{odena2016deconvolution}
Odena, A.; Dumoulin, V.; and Olah, C. 2016.
\newblock Deconvolution and Checkerboard Artifacts.
\newblock \emph{Distill}.

\bibitem[{Park et~al.(2022)Park, Vien, Kim, Koh, and Lee}]{cyclic}
Park, H.; Vien, A.~G.; Kim, H.; Koh, Y.~J.; and Lee, C. 2022.
\newblock Unpaired screen-shot image demoir{\'e}ing with cyclic moir{\'e} learning.
\newblock \emph{IEEE Access}, 10: 16254--16268.

\bibitem[{Porter and Duff(1984)}]{Porter_Duff_1984}
Porter, T.; and Duff, T. 1984.
\newblock Compositing digital images.
\newblock \emph{ACM SIGGRAPH Computer Graphics}, 253–259.

\bibitem[{Rombach et~al.(2022)Rombach, Blattmann, Lorenz, Esser, and Ommer}]{Rombach2022LDM}
Rombach, R.; Blattmann, A.; Lorenz, D.; Esser, P.; and Ommer, B. 2022.
\newblock High-Resolution Image Synthesis with Latent Diffusion Models.
\newblock In \emph{2022 IEEE/CVF Conference on Computer Vision and Pattern Recognition (CVPR)}.

\bibitem[{Ronneberger, Fischer, and Brox(2015)}]{ronneberger2015u}
Ronneberger, O.; Fischer, P.; and Brox, T. 2015.
\newblock U-net: Convolutional networks for biomedical image segmentation.
\newblock In \emph{International Conference on Medical image computing and computer-assisted intervention}, 234--241. Springer.

\bibitem[{Simonyan and Zisserman(2014)}]{simonyan2014very}
Simonyan, K.; and Zisserman, A. 2014.
\newblock Very deep convolutional networks for large-scale image recognition.
\newblock \emph{arXiv preprint arXiv:1409.1556}.

\bibitem[{Song, Meng, and Ermon(2022)}]{song2022ddim}
Song, J.; Meng, C.; and Ermon, S. 2022.
\newblock Denoising Diffusion Implicit Models.
\newblock arXiv:2010.02502.

\bibitem[{Sun, Li, and Sun(2014)}]{sun2014scanned}
Sun, B.; Li, S.; and Sun, J. 2014.
\newblock Scanned image descreening with image redundancy and adaptive filtering.
\newblock \emph{IEEE transactions on image processing}, 23(8): 3698--3710.

\bibitem[{Sun, Yu, and Wang(2018)}]{sun2018moire}
Sun, Y.; Yu, Y.; and Wang, W. 2018.
\newblock Moir{\'e} photo restoration using multiresolution convolutional neural networks.
\newblock \emph{IEEE Transactions on Image Processing}, 27(8): 4160--4172.

\bibitem[{Ulyanov, Vedaldi, and Lempitsky(2016)}]{Ulyanov2016IN}
Ulyanov, D.; Vedaldi, A.; and Lempitsky, V. 2016.
\newblock Instance Normalization: The Missing Ingredient for Fast Stylization.
\newblock \emph{arXiv: Computer Vision and Pattern Recognition,arXiv: Computer Vision and Pattern Recognition}.

\bibitem[{Wang et~al.(2023{\natexlab{a}})Wang, He, Wu, Wan, Shi, and Duan}]{wang2023coarse}
Wang, C.; He, B.; Wu, S.; Wan, R.; Shi, B.; and Duan, L.-Y. 2023{\natexlab{a}}.
\newblock Coarse-to-fine Disentangling Demoir{\'e}ing Framework for Recaptured Screen Images.
\newblock \emph{IEEE Transactions on Pattern Analysis and Machine Intelligence}.

\bibitem[{Wang et~al.(2019)Wang, Chen, Xu, Liu, Loy, and Lin}]{wang2019carafe}
Wang, J.; Chen, K.; Xu, R.; Liu, Z.; Loy, C.~C.; and Lin, D. 2019.
\newblock CARAFE: Content-Aware ReAssembly of FEatures.
\newblock arXiv:1905.02188.

\bibitem[{Wang et~al.(2023{\natexlab{b}})Wang, Zhang, Shen, Luo, Stenger, and Lu}]{wang2023uhdlol4k}
Wang, T.; Zhang, K.; Shen, T.; Luo, W.; Stenger, B.; and Lu, T. 2023{\natexlab{b}}.
\newblock Ultra-high-definition low-light image enhancement: A benchmark and transformer-based method.
\newblock In \emph{Proceedings of the AAAI Conference on Artificial Intelligence}, volume~37, 2654--2662.

\bibitem[{Wang et~al.(2021)Wang, Xie, Dong, and Shan}]{wang2021real}
Wang, X.; Xie, L.; Dong, C.; and Shan, Y. 2021.
\newblock Real-ESRGAN: Training Real-World Blind Super-Resolution with Pure Synthetic Data.
\newblock In \emph{Proceedings of the IEEE/CVF International Conference on Computer Vision}, 1905--1914.

\bibitem[{Wang et~al.(2004)Wang, Bovik, Sheikh, and Simoncelli}]{wang2004image}
Wang, Z.; Bovik, A.~C.; Sheikh, H.~R.; and Simoncelli, E.~P. 2004.
\newblock Image quality assessment: from error visibility to structural similarity.
\newblock \emph{IEEE transactions on image processing}, 13(4): 600--612.

\bibitem[{Wang et~al.(2022)Wang, Cun, Bao, Zhou, Liu, and Li}]{Wang2022Uformer}
Wang, Z.; Cun, X.; Bao, J.; Zhou, W.; Liu, J.; and Li, H. 2022.
\newblock Uformer: A General U-Shaped Transformer for Image Restoration.
\newblock In \emph{2022 IEEE/CVF Conference on Computer Vision and Pattern Recognition (CVPR)}.

\bibitem[{Xing and Egiazarian(2021)}]{xing2021end}
Xing, W.; and Egiazarian, K. 2021.
\newblock End-to-End Learning for Joint Image Demosaicing, Denoising and Super-Resolution.
\newblock In \emph{Proceedings of the IEEE/CVF Conference on Computer Vision and Pattern Recognition}, 3507--3516.

\bibitem[{Yang et~al.(2023)Yang, Yang, Ke, Chen, Grzegorzek, and See}]{yang2023doing}
Yang, C.; Yang, Z.; Ke, Y.; Chen, T.; Grzegorzek, M.; and See, J. 2023.
\newblock Doing More With Moir{\'e} Pattern Detection in Digital Photos.
\newblock \emph{IEEE Transactions on Image Processing}, 32: 694--708.

\bibitem[{Yang et~al.(2017{\natexlab{a}})Yang, Liu, Yue, Fu, Hou, and Wu}]{yang2017textured}
Yang, J.; Liu, F.; Yue, H.; Fu, X.; Hou, C.; and Wu, F. 2017{\natexlab{a}}.
\newblock Textured image demoir{\'e}ing via signal decomposition and guided filtering.
\newblock \emph{IEEE Transactions on Image Processing}, 26(7): 3528--3541.

\bibitem[{Yang et~al.(2017{\natexlab{b}})Yang, Zhang, Cai, and Li}]{yang2017demoireing}
Yang, J.; Zhang, X.; Cai, C.; and Li, K. 2017{\natexlab{b}}.
\newblock Demoir{\'e}ing for screen-shot images with multi-channel layer decomposition.
\newblock In \emph{2017 IEEE Visual Communications and Image Processing (VCIP)}, 1--4. IEEE.

\bibitem[{Yu et~al.(2021)Yu, Li, Koh, Zhang, Pang, Qin, Ku, Xu, Baldridge, and Wu}]{Yu2021Vector}
Yu, J.; Li, X.; Koh, J.; Zhang, H.; Pang, R.; Qin, J.; Ku, A.; Xu, Y.; Baldridge, J.; and Wu, Y. 2021.
\newblock Vector-quantized Image Modeling with Improved VQGAN.
\newblock \emph{Cornell University - arXiv,Cornell University - arXiv}.

\bibitem[{Yu et~al.(2022)Yu, Dai, Li, Ma, Shen, Li, and Qi}]{yu2022towards}
Yu, X.; Dai, P.; Li, W.; Ma, L.; Shen, J.; Li, J.; and Qi, X. 2022.
\newblock Towards efficient and scale-robust ultra-high-definition image demoir{\'e}ing.
\newblock In \emph{European Conference on Computer Vision}, 646--662. Springer.

\bibitem[{Yuan et~al.(2019)Yuan, Timofte, Slabaugh, Leonardis, Zheng, Ye, Tian, Chen, Cheng, Fu et~al.}]{yuan2019aim}
Yuan, S.; Timofte, R.; Slabaugh, G.; Leonardis, A.; Zheng, B.; Ye, X.; Tian, X.; Chen, Y.; Cheng, X.; Fu, Z.; et~al. 2019.
\newblock Aim 2019 challenge on image demoireing: Methods and results.
\newblock In \emph{2019 IEEE/CVF International Conference on Computer Vision Workshop (ICCVW)}, 3534--3545. IEEE.

\bibitem[{Yue et~al.(2022)Yue, Cheng, Mao, Cao, and Yang}]{yue2022recaptured}
Yue, H.; Cheng, Y.; Mao, Y.; Cao, C.; and Yang, J. 2022.
\newblock Recaptured screen image demoir{\'e}ing in raw domain.
\newblock \emph{IEEE Transactions on Multimedia}.

\bibitem[{Zhang et~al.(2023)Zhang, Zhu, Yan, Sun, and Zhang}]{zhang2023all}
Zhang, C.; Zhu, Y.; Yan, Q.; Sun, J.; and Zhang, Y. 2023.
\newblock All-in-one multi-degradation image restoration network via hierarchical degradation representation.
\newblock In \emph{Proceedings of the 31st ACM International Conference on Multimedia}, 2285--2293.

\bibitem[{Zhang et~al.(2018)Zhang, Isola, Efros, Shechtman, and Wang}]{zhang2018unreasonable}
Zhang, R.; Isola, P.; Efros, A.~A.; Shechtman, E.; and Wang, O. 2018.
\newblock The unreasonable effectiveness of deep features as a perceptual metric.
\newblock In \emph{Proceedings of the IEEE conference on computer vision and pattern recognition}, 586--595.

\bibitem[{Zheng et~al.(2020)Zheng, Yuan, Slabaugh, and Leonardis}]{zheng2020image}
Zheng, B.; Yuan, S.; Slabaugh, G.; and Leonardis, A. 2020.
\newblock Image demoireing with learnable bandpass filters.
\newblock In \emph{Proceedings of the IEEE/CVF Conference on Computer Vision and Pattern Recognition}, 3636--3645.

\bibitem[{Zheng et~al.(2021)Zheng, Yuan, Yan, Tian, Zhang, Sun, Liu, Leonardis, and Slabaugh}]{zheng2021learning}
Zheng, B.; Yuan, S.; Yan, C.; Tian, X.; Zhang, J.; Sun, Y.; Liu, L.; Leonardis, A.; and Slabaugh, G. 2021.
\newblock Learning frequency domain priors for image demoireing.
\newblock \emph{IEEE Transactions on Pattern Analysis and Machine Intelligence}, 44(11): 7705--7717.

\bibitem[{Zhong et~al.(2024)Zhong, Zhou, Zhang, Chao, and Ji}]{undem}
Zhong, Y.; Zhou, Y.; Zhang, Y.; Chao, F.; and Ji, R. 2024.
\newblock Learning Image Demoireing from Unpaired Real Data.
\newblock \emph{arXiv preprint arXiv:2401.02719 (AAAI2024)}.

\bibitem[{Zhou et~al.(2021)Zhou, Yang, Qiao, and Xiang}]{Zhou2021Mixstyle}
Zhou, K.; Yang, Y.; Qiao, Y.; and Xiang, T. 2021.
\newblock MixStyle Neural Networks for Domain Generalization and Adaptation.
\newblock \emph{Cornell University - arXiv,Cornell University - arXiv}.

\bibitem[{Zhu et~al.(2023)Zhu, Zhang, Liang, Cao, Wen, Timofte, and Van~Gool}]{zhu2023denoising}
Zhu, Y.; Zhang, K.; Liang, J.; Cao, J.; Wen, B.; Timofte, R.; and Van~Gool, L. 2023.
\newblock Denoising Diffusion Models for Plug-and-Play Image Restoration.
\newblock In \emph{Proceedings of the IEEE/CVF Conference on Computer Vision and Pattern Recognition}, 1219--1229.

\end{thebibliography}

\newpage
\clearpage
\appendix
\setcounter{secnumdepth}{2} 
\section*{Technical Appendix}
\label{sec:sup_outline}

This document supplements the main body of our paper with additional details, discussions, and results. In Section~\ref{sec:sup_dataset}, we present more details of the Moiré Pattern Dataset collection, including a brief analysis of various previously overlooked factors affecting moiré pattern diversity. In Section~\ref{sec:sup_method}, we will provide a detailed explanation of the two stages involved in implementing UniDemoiré: Moiré Pattern Generator and Moiré Image Synthesis. In Section~\ref{sec:sup_esperiments}, we provide more implementation details of experiments and show more qualitative results. Furthermore, as shown in Section~\ref{sec:sup_ablation}, we performed additional ablation experiments on the blending strategy in the Moiré Image Blending (MIB) module and the design of the upsampling block and the loss function in the Tone Refinement Network (TRN).

\section{Dataset Capture and Analysis}
\label{sec:sup_dataset}

\begin{table*}[t]

\centering
\vspace{-8pt}
\scalebox{0.95}
{      
\begin{tabular}{c|ccccc}
\toprule[1.25pt]    
Mobile Phone & Camera & CMOS & MegaPixel ($\text{MP}$) & Optical format ($\text{Inches}$) & Pixel Size ($\mu m$)   \\
\midrule
\multirow{1}{*}{iPhone 12}     & Main      & SONY   IMX503            & 12   & 1/2.55 & 1.40  \\
\multirow{1}{*}{iPhone 13}     & Main      & SONY IMX603              & 12   & 1/1.88 & 1.70  \\
\multirow{1}{*}{Honor 90}      & Main      & ISOCELL HP3              & 200  & 1/1.40 & 0.56  \\
\multirow{1}{*}{Xiaomi 10s}    & Main      & ISOCELL HMX              & 108  & 1/1.33 & 0.80  \\
\midrule
\multirow{2}{*}{iPhone 12 Pro} & Main      & SONY IMX503              & 12   & 1/2.55 & 1.40  \\
                               & Telephoto & SONY IMX613 (2x zoom)    & 12.2 & 1/3.40 & 1.00  \\
\midrule
\multirow{2}{*}{iPhone 15 Pro} & Main      & SONY IMX803              & 48   & 1/1.28 & 1.22  \\
                               & Telephoto & SONY IMX713 (2x/4x zoom) & 12   & 1/3.40 & 1.00  \\

\bottomrule[1.25pt]  
\end{tabular}
}
\caption{The mobile phone we apply to get the moiré patterns}
\label{tab:Datasets-Phone}
\end{table*}

\begin{table*}[t]

\centering
\vspace{-8pt}
\scalebox{1.0}
{      
\begin{tabular}{c|ccccc}
\toprule[1.25pt]
Digital Screen    & Size ($Inches$) & Panel type        & Resolution         & Brightness ($cd/m^2$) & Coating \\
\midrule
DELL D2720DS      & 27              & IPS(LED)          & 2560 $\times$ 1440 & 350                   & Matte      \\
Macbook Air 2022  & 13.3            & IPS(LED)          & 2560 $\times$ 1600 & 500                   & Glossy     \\
AOC 27G2G8        & 27              & IPS(W-LED)        & 2560 $\times$ 1440 & 250                   & Matte      \\
Philips 27E1N5500 & 27              & IPS(LED)          & 2560 $\times$ 1440 & 300                   & Matte      \\
Xiaomi C34WQBA-RG & 34              & Curved SVA(W-LED) & 3440 $\times$ 1440 & 300                   & Matte      \\
ViewSonic VX2771-4K-HD & 27         & IPS(LED)          & 3840 $\times$ 2160 & 350                   & Matte      \\
\bottomrule[1.25pt]
\end{tabular}
}
\caption{The screen we apply to get the moiré patterns}
\label{tab:Datasets-Screen}
\end{table*}

In this section, we first present a brief introduction of various previously overlooked factors of devices that affect moiré pattern diversity. Then, we provide more details about our capture settings.

\subsection{The Impact of Device on Moiré Pattern Diversity}


Previous studies~\cite{yu2022towards, yang2023doing} have indicated that the geometric correlation between the screen and the camera significantly influences the features of the moiré pattern. However, such studies have overlooked that some aspects of the camera and the screen can also impact the moiré pattern.

For cameras, the two most critical factors affecting the moiré pattern are the CMOS and the lens used. The pixel density of a CMOS sensor (i.e., the number of pixels per unit area) determines its maximum sampling frequency, also known as the Nyquist frequency. The higher the pixel density, the higher the sampling frequency of the sensor and the higher the frequency of the signal that can be sampled, resulting in a higher frequency of moiré produced by the aliasing effect, which impacts the moiré pattern. In addition, the lens's focal length also affects the formation of moiré. In cell phone photography, lenses with shorter focal lengths (e.g., wide lenses/main camera lenses) usually have wider angles of view and can capture more of the scene content. Lenses with longer focal lengths (such as telephoto or telescopic lenses), on the other hand, offer a narrower angle of view and greater magnification for capturing distant details. When the screen is photographed with lenses of different focal lengths, the relative positional relationship between the pixels on the sensor and the pixels on the screen changes, which may cause the moiré pattern to appear or disappear.

\begin{figure}[t]
  \centering
    \includegraphics[width=1.0\linewidth]{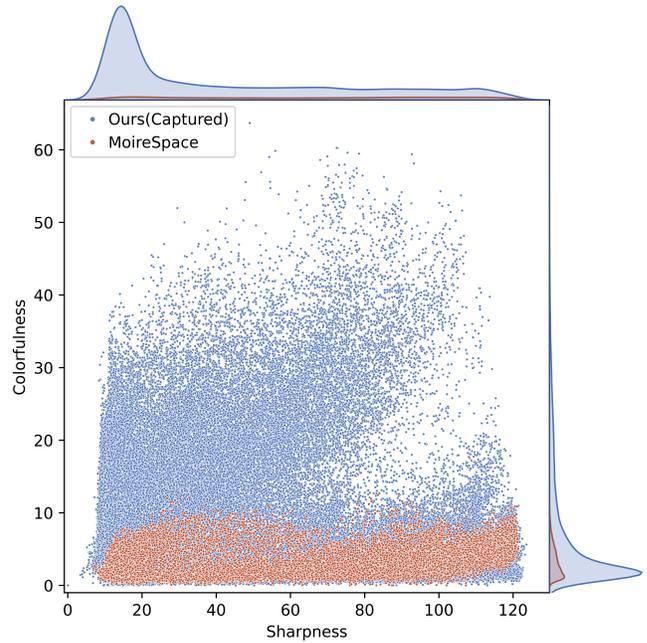}
  \caption{Comparison of sharpness and colorfulness between our Moiré Pattern Dataset and MoireSpace.} 
  \label{fig:distribution}
\end{figure}

\begin{figure*}[!p]
  \centering
    \includegraphics[width=1.0\linewidth]{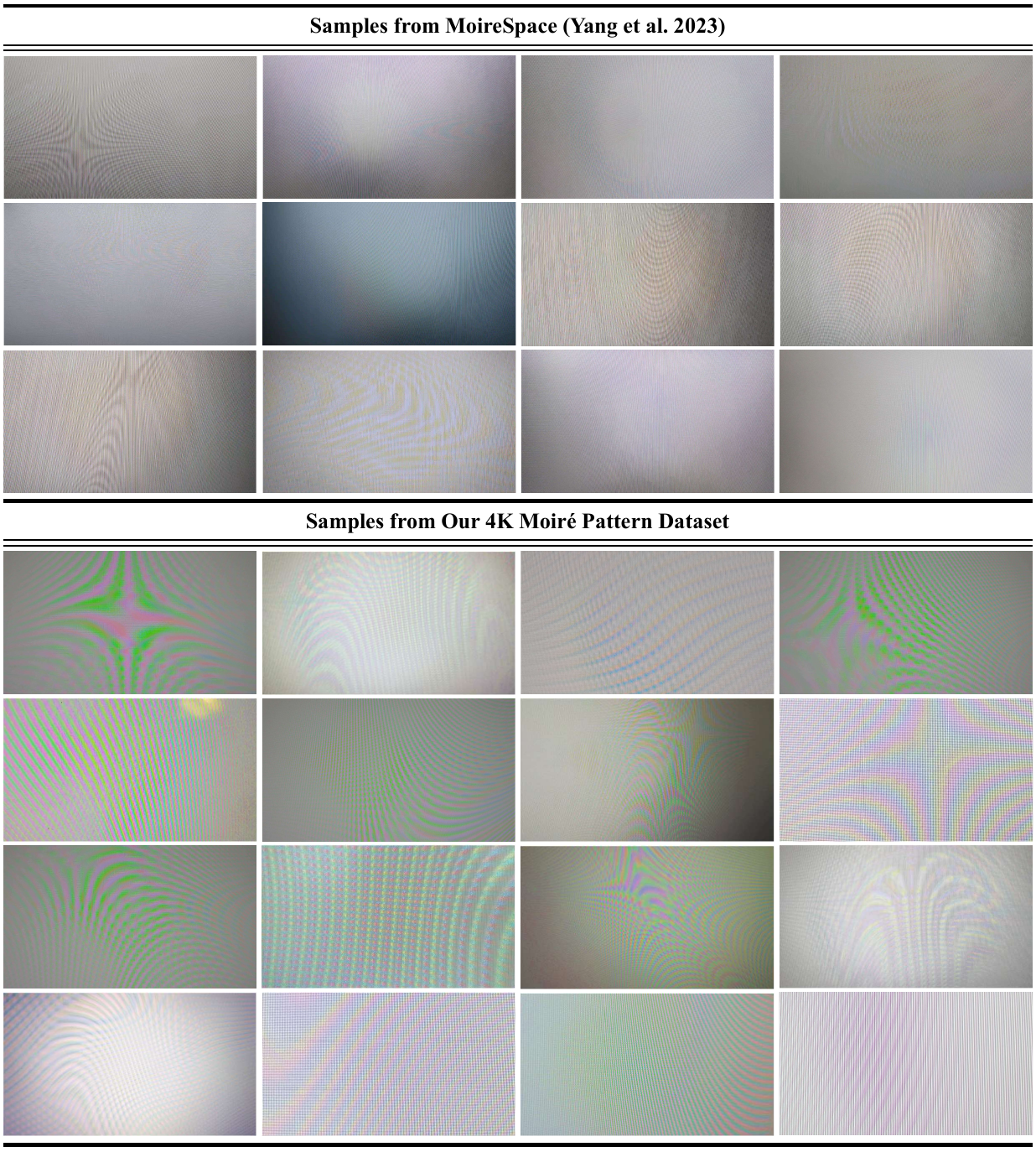}
  \caption{Samples from MoireSpace~\cite{yang2023doing} and our 4K Moiré Pattern Dataset.}
  \label{fig:capture_data}
\end{figure*}

Furthermore, the layout and the distance of pixels dots in the panel used can also significantly impact the formation of moiré on the display screen. The frequency of detail that a screen can display depends on how the pixel dots are arranged. Various arrangements result in distinct frequencies of detail, which impacts the formation of moiré patterns. The distance between pixel dots on the screen then affects the shooting distance. Larger pixel dot spacing will make the distance at which the moiré is formed to be photographed farther away. Conversely, the smaller the distance between pixel dots, the closer the distance needed to photograph the molded moiré.

\subsection{More Details about Capture Settings}
Based on the above analysis, we take screen images through different camera viewpoints to generate diverse moiré patterns. Specifically, we apply six mobile phones and six digital screens, as shown in Table~\ref{tab:Datasets-Phone} and ~\ref{tab:Datasets-Screen} ($\text{6} \times \text{6} = \text{36}$ combinations). 
Figure~\ref{fig:distribution} presents a scatter plot comparison between our 4K Moiré Pattern Dataset and the MoireSpace dataset in terms of texture definition and color vibrancy, where each data point represents an individual sample. As illustrated in the visualization, our captured moiré patterns demonstrate a more comprehensive coverage in both the sharpness of textural details and the chromatic saturation range. This comparative analysis is further substantiated by the representative samples juxtaposed in Figure~\ref{fig:capture_data}, where side-by-side comparisons visually confirm that our dataset encompasses a broader diversity of morphological textures and a wider gamut of color expressions.

\subsubsection{Mobile Phones} 
We chose six mobile phones with varying camera specifications to capture diverse moiré patterns, as shown in Table~\ref{tab:Datasets-Phone}. Our selection criteria included the camera type, CMOS category, and number of megapixels. For the regular main camera with moderate resolution, we picked the iPhone 12 and iPhone 13. For electronic zooming at 2x and 3x, we selected the Honor 90 and Xiaomi 10s, which have high pixels. Additionally, we picked two iPhone 12 Pro and iPhone 15 Pro models with different CMOS specifically for telephoto lenses. These models use the telephoto lens for optical zoom at fixed magnifications of 2x and 3x.

\subsubsection{Display Screens} 
To capture a wider variety of moiré patterns in different forms, we selected display screens based on size, panel type, and resolution guidelines to maximize pixel point layouts and spacing on the screen. As shown in Table~\ref{tab:Datasets-Screen}, we have selected three 27-inch IPS panel LED matte screen monitors with a 2K resolution - DELL D2720DS, AOC 27G2G8, and Philips 27E1N5500. This specification is the most common among the available options. The AOC 27G2G8 is a W-LED monitor with an RGBW pixel layout. This IPS screen has white sub-pixels in addition to the standard RGB arrangement, creating a more varied pixel point layout. To capture the moiré pattern on the glossy display, we opted for a 13.3-inch IPS panel with a 2K resolution MacBook Air notebook. Finally, we selected two high-resolution displays: the Xiaomi C34WQBA-RG and the ViewSonic VX2771-4K-HD. These displays were explicitly chosen to capture moiré patterns with smaller pixel dot spacing. The Xiaomi C34WQBA-RG is a 34-inch curved display with an SVA panel and W-LED technology. It boasts a 3K resolution. On the other hand, the ViewSonic VX2771-4K-HD is a 27-inch matte screen display with an IPS panel and LED technology. It offers a standard 4K resolution.

\section{Further details of our Method}
\label{sec:sup_method}

This section will showcase the details of the implementation of our UniDemoiré's Moiré Pattern Generator stage and Moiré Image Synthesis stage.

\subsection{Moiré Pattern Generator}
The visualization of moiré pattern patches generated using the Moiré Pattern Generator(MPG) is shown in Figure~\ref{fig:generated_data}. The details of the data preprocessing and networking implementations of MPG are described below.

\begin{algorithm}[!t]
\caption{Data Preprocessing in MPG.}
\label{sup:alg_data_preprocessing} 
    \DontPrintSemicolon
    \LinesNumbered
    \KwIn{4K Moiré Pattern Dataset $\mathcal{D}_{mp}$, patch size $(w, h)$.}
    \KwOut{Selected moiré pattern patch $I_{mp}$.}
    \While{True}{
        Randomly select a 4K moiré pattern $\mathcal{I} \in \mathcal{D}_{mp}$. \;
        \For{i = $1$ to $n$}{
            \SetKwProg{Fn}{1. Multi-Scale Cropping}{:}{}
            \Fn{}{
                Randomly select probability $p_1$, $p_2$. \;
                \uIf{$p_1 \leq 50\%$}{
                    $I_{mp} \leftarrow \text{Random Crop}(\mathcal{I}, w, h)$. \;
                }\uElseIf{$p_2 \leq 33.33\%$}{
                    $\mathcal{I} \leftarrow \text{Resize}(\mathcal{I}, 2560, 1440)$, \; 
                    $I_{mp} \leftarrow \text{Random Crop}(\mathcal{I}, w, h)$. \;
                }\uElseIf{$33.33\% \space \textless \space p_2 \leq 66.66\%$}{
                    $\mathcal{I} \leftarrow \text{Resize}(\mathcal{I}, 1920, 1080)$, \;
                    $I_{mp} \leftarrow \text{Random Crop}(\mathcal{I}, w, h)$. \;
                }\lElse{
                    $I_{mp} \leftarrow \text{Resize}(\mathcal{I}, w, h)$.
                }
            }
    
            \SetKwProg{Fn}{2. Sharpness-Colorfulness selection}{:}{}
            \Fn{}{
                $G_{mp} \leftarrow \text{RGB\_to\_Gray}(I_{mp})$, \;
                $L_{mp}, A_{mp}, B_{mp} \leftarrow \text{RGB\_to\_LAB}(I_{mp})$, \;
                $\text{Sharpness} \leftarrow \sigma(\mathcal{F} \ast G_{mp})$, \;
                $\text{Colorfulness} \leftarrow \sqrt{\sigma(A_{mp})^2 + \sigma(B_{mp})^2}$, \;
                \uIf{$\text{Sharpness} \geq \delta_s$ and $\text{Colorfulness} \geq \delta_c$}{
                    \KwRet $I_{mp}$. \;
                }
            }
        }    
    }
\end{algorithm}

\subsubsection{The implementation details of data preprocessing} The details of our data preprocessing method in the Moiré Pattern Generator(MPG) are described in Algorithm~\ref{sup:alg_data_preprocessing}. In Multi-Scale Cropping, ``$\text{Random Crop}(\mathcal{I},w,h)$'' means to randomly crop a patch of size $w\times h$ from $\mathcal{I}$, and ``$\text{Resize}(\mathcal{I},w,h)$'' means to resize the width and height of $\mathcal{I}$ to $w$ and $h$ directly. In Sharpness-Colorfulness selection, ``$\text{RGB\_to\_Gray}(I_{mp})$'' refers to convert $I_{mp}$ to grayscale image $G_{mp}$, while ``$\text{RGB\_to\_LAB}(I_{mp})$'' refers to convert $I_{mp}$ to LAB space and retrieve the corresponding channel matrices $L_{mp}$, $A_{mp}$, and $B_{mp}$ respectively. Moreover, ``$\mathcal{F} \ast G_{mp}$'' denotes the convolution operation on the grayscale image $G_{mp}$ using the Laplace edge detection operator $\mathcal{F}$. 
In the actual training process of MPG, we specify set $n$ to 3 and ($w$, $h$) to (768,768), while $\delta_s$ and $\delta_c$ are set to 15 and 2, respectively.

\subsubsection{The implementation details of Latent Diffusion Model} 
We utilize the Latent Diffusion Model(LDM)~\cite{Rombach2022LDM} as the network component of the Moiré Pattern Generator. 

\begin{figure*}[!p]
  \centering
    \includegraphics[width=1.0\linewidth]{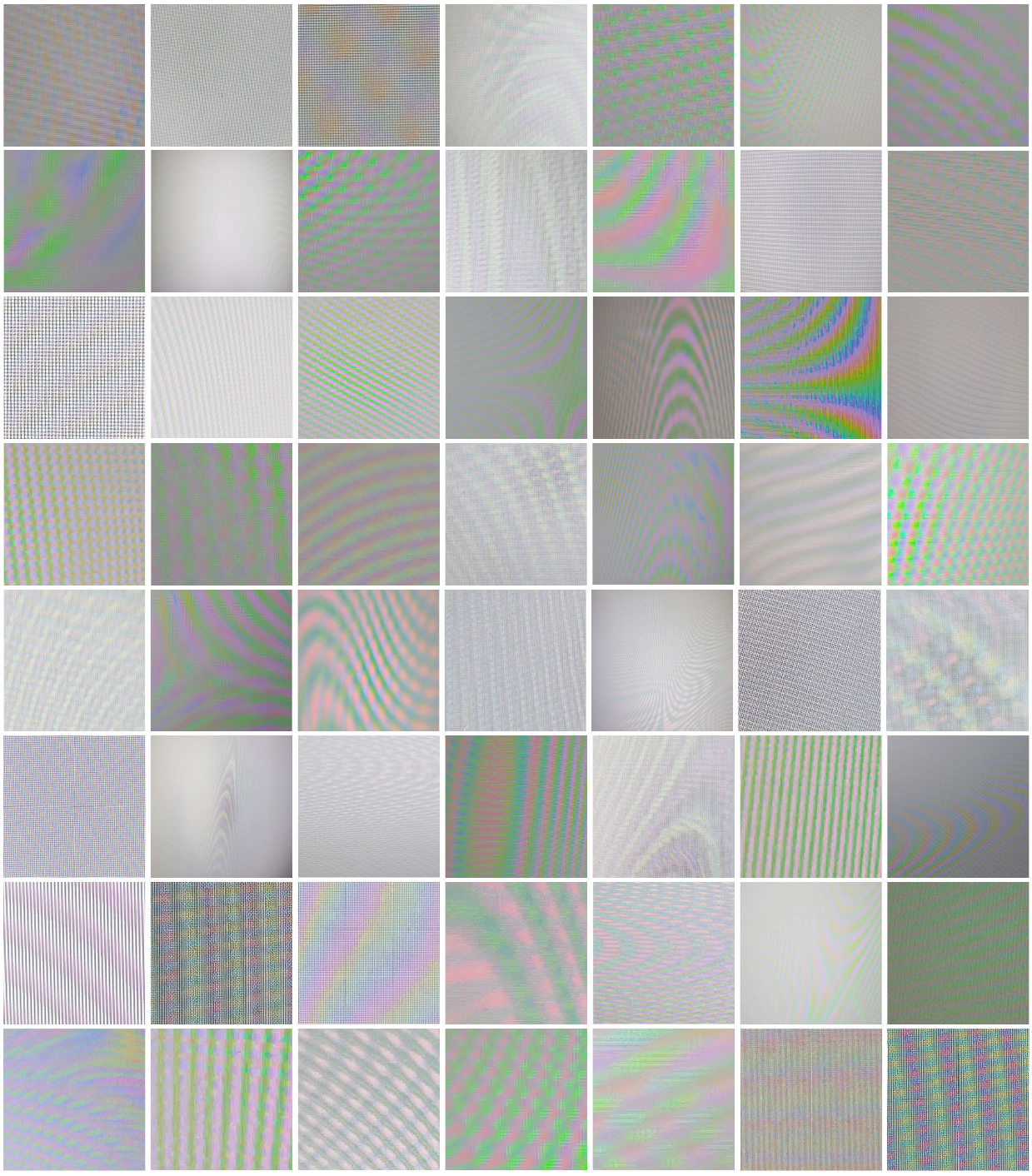}
  \caption{Visualization of sampled patches using our Moiré Pattern Generator.}
  \label{fig:generated_data}
\end{figure*}

\begin{figure*}[!t]
  \centering
    \includegraphics[width=1.0\linewidth]{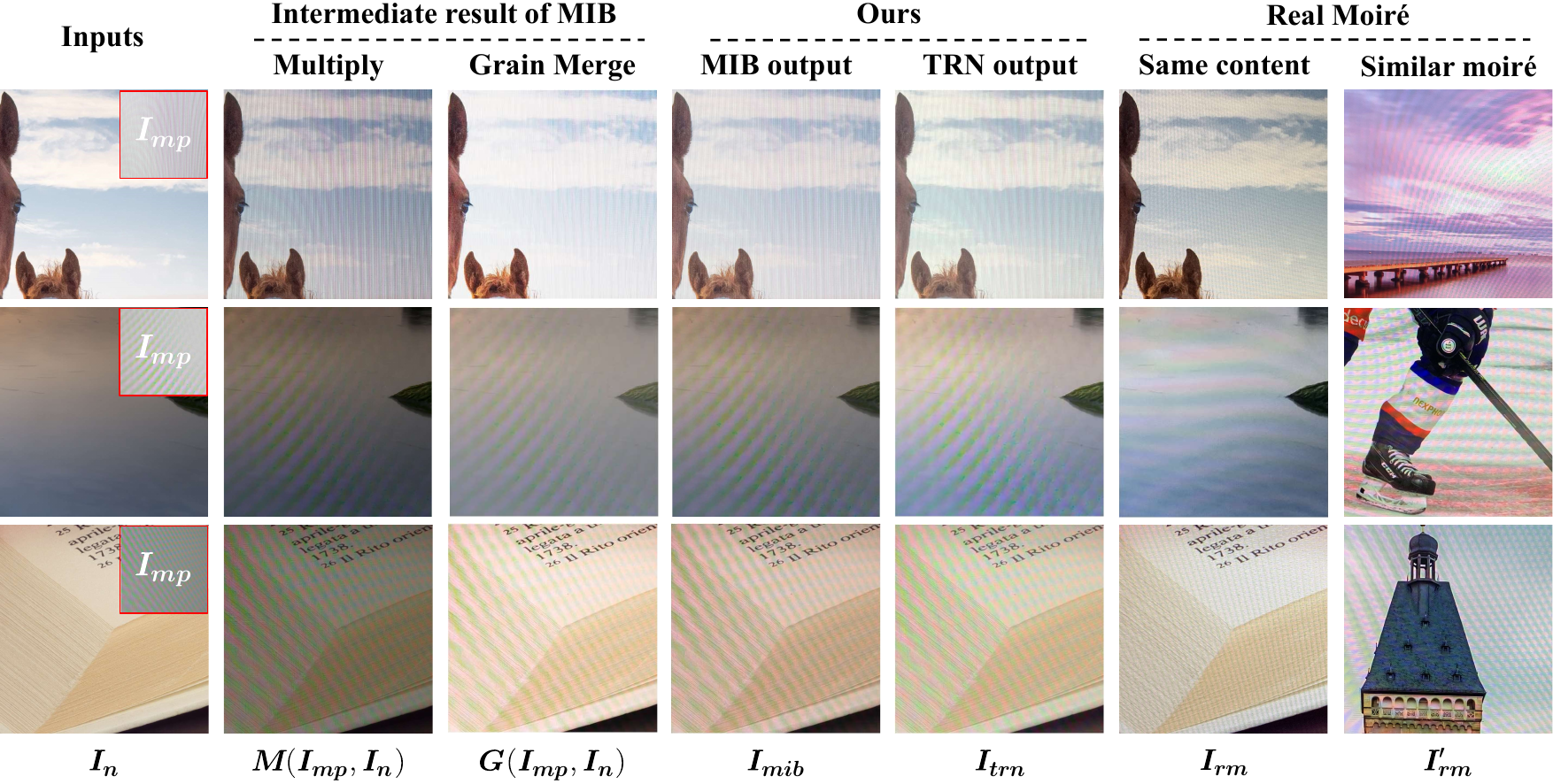}
  \caption{Visualization of our intermediate synthetic results. The final synthesis of $I_{trn}$ best resembles the real moiré images in contrast and brightness distortions.} 
  \label{fig:synthesis_result_sup}
\end{figure*}

Firstly, we have trained our autoencoder model for moiré patterns according to the method described in ~\cite{Rombach2022LDM}.
Specifically, given a moiré pattern patch $I_{mp} \in \mathrm{R}^{w \times h \times 3}$ that has gone through Multi-Scale Cropping and Sharpness-Colorfulness selection, we utilize an Encoder $\mathcal{E}$ to convert $I_{mp}$ to the latent space $z=\mathcal{E}(I _{mp})$ through multiple downsampling blocks. Simultaneously, we expect the corresponding Decoder $\mathcal{D}$ to reconstruct the moiré pattern from the latent space variable $z$ : $I_{mp}=\mathcal{D}(z)=\mathcal{D}(\mathcal{E}(I_{mp}))$ by using the same upsampling factor. Note that the overall downsampling factor is denoted as $f=h/h_0=w/w_0$, where  $h_0$ and $w_0$  are hyperparameters chosen to ensure that $f$ is precisely $2^m$, with $m \in \mathrm{N}$.
Our loss function is a combination of a perceptual loss function $\mathcal{L}_{rec}$~\cite{zhang2018unreasonable} and patch-based adversarial targets $\mathcal{L}_{adv}$~\cite{Dosovitskiy2016Generating, Esser2021Taming, Yu2021Vector}, along with a KL-reg regularization term $\mathcal{L}_{reg}$ where the patch-based discriminator $D_{\psi}$ we used is optimized to differentiate between the original moiré pattern $I_{mp}$ and the reconstructed moiré pattern $\mathcal{D}(\mathcal{E}(I_{mp}))$. 
The full objective to train the autoencoder $(\mathcal{E},\mathcal{D})$ is:
\begin{equation}
\begin{split}
&\mathcal{L}=\min_{\mathcal{E}, \mathcal{D}} \max_{\psi}(\mathcal{L}_{rec}(I_{mp}, \mathcal{D}(\mathcal{E}(I_{mp})))+\log D_\psi(I_{mp}) \\
&-\mathcal{L}_{adv}(\mathcal{D}(\mathcal{E}(I_{mp})))+\mathcal{L}_{reg}(I_{mp} ; \mathcal{E}, \mathcal{D}))
\end{split}
\end{equation}

\begin{figure*}[!t]
  \centering
    \includegraphics[width=1.0\linewidth]{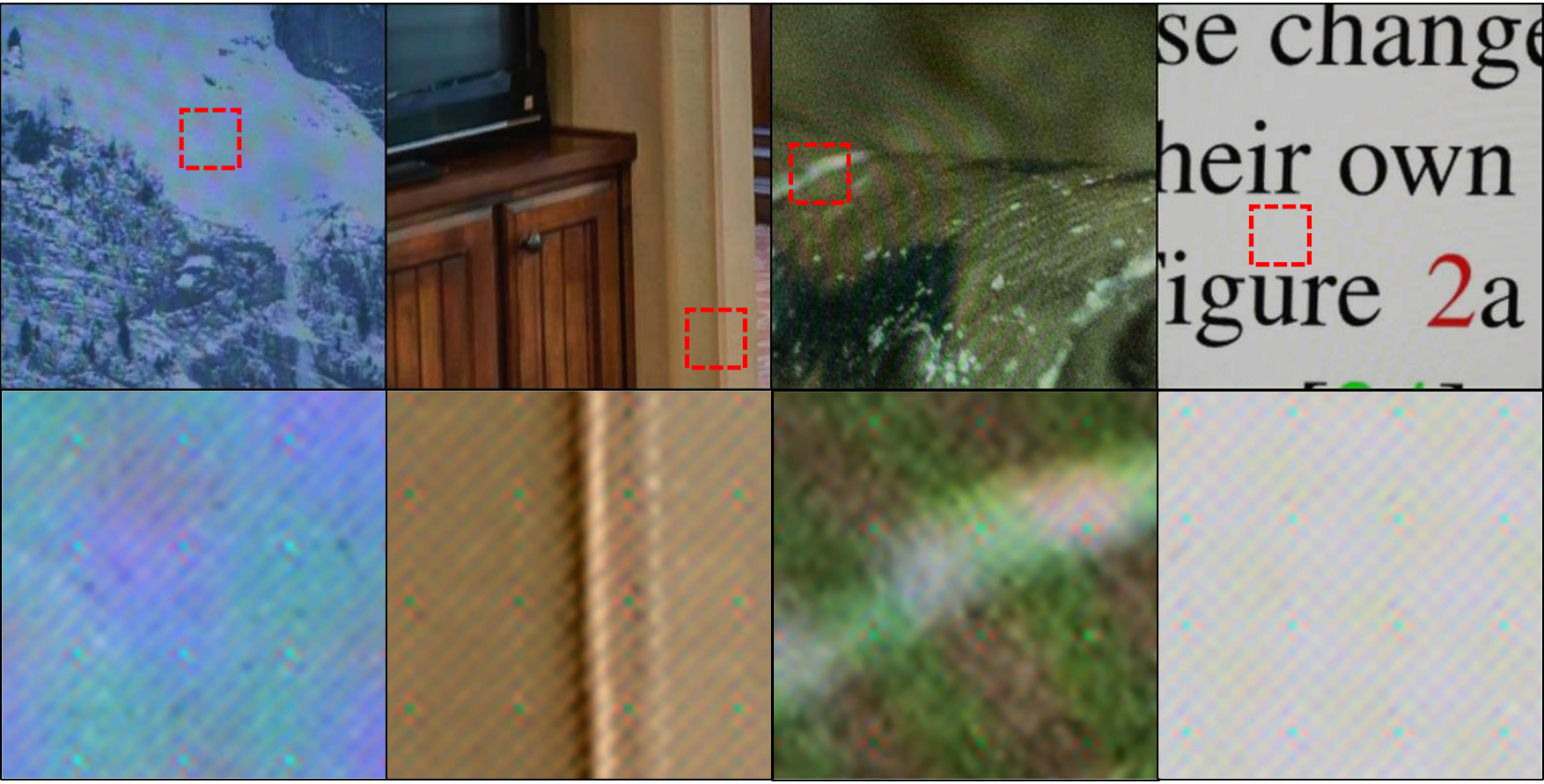}
  \caption{Examples of the ``Checkerboard Artifacts'' that occur in the $I_{trn}$ when upsampling with Uformer’s transpose convolution~\cite{Wang2022Uformer}.} 
  \label{fig:checkerboard_artifacts}
\end{figure*}

The network structure of our $\mathcal{E}$ and $\mathcal{D}$ are the same as the autoencoder in ~\cite{Rombach2022LDM}.
To compress the moiré pattern as much as possible, we use the downsampling factor $f=\text{32}$ and the number of hidden-space channels 64, which gives the latent variable $z$ a dimension of 64 $\times$ 64 $\times$ 24. 
We adopted 6 downsampling/upsampling blocks in $\mathcal{E}$ and $\mathcal{D}$. Each downsampling/upsampling block contains two layers of ResBlock as well as one layer of multi-head self-attention block, and the list of channel scaling multipliers is [1,1,2,2,4,4]. We trained the autoencoder on 8 NVIDIA A40 GPUs, with a batch size of 2 on each GPU and a learning rate of 4.5e-6. 
Both the autoencoder ($\mathcal{E}$ and $\mathcal{D}$) and the discriminator ($D_{\psi}$) in the loss functions are optimized by Adam~\cite{kingma2014adam} with $\beta_1$ = 0.5 and $\beta_2$ = 0.9.
In total, we trained the autoencoder for 35 Epochs.

Subsequently, we adopt the diffusion model to modify the complex distribution $p(z)$ obeyed by the latent variable $z$ after the $\mathcal{E}$ transformation with the objective function:
\begin{equation}
\mathcal{L}=\mathbb{E}_{\mathcal{E}\left(I_{m p}\right), \epsilon, t}\left[\| \epsilon-\epsilon_\theta\left(\alpha_t \mathcal{E}\left(I_{m p}\right)+\sigma_t \epsilon, t\right) \|_1\right].
\end{equation}
where $\epsilon\sim\mathcal{N}(0, \mathbb{I})$ is the variable sampled by the standard Gaussian distribution, and $\epsilon_\theta$ is the noisy prediction network parameterized by $\theta$, where we implemented it by using UNet~\cite{ronneberger2015u} which integrates the time-step conditioning variable $t$. The $\alpha_t$ is the value at step $t$ of the signal-to-noise ratio.

The network structures of the diffusion model in latent space are the same as those of the unconditional model in LDM~\cite{Rombach2022LDM}. For the UNet model $\epsilon_\theta$, we set the number of channels to 192, and the encoder and decoder of the UNet contain 4 downsampling/upsampling blocks, and the structure of those blocks are kept the same in the autoencoder. The list of channel scaling multipliers is [1,2,4,8]. 
The model is trained on 8 NVIDIA A40 GPUs for 50 epochs and optimized by AdamW~\cite{loshchilov2019adamw} with $\beta_1$ = 0.9 and $\beta_2$ = 0.999.
The batch size on each GPU is set to 2, and the learning rate is initially set to 1e-4 and scheduled by linear warmup on the first 10,000 steps. The total learning rate complies with the linear multiplication in the LDM~\cite{Rombach2022LDM} based on the number of GPUs and the batch size.
In the training stage, we set the diffusion steps to 1000 and utilized the linear noise schedule to add noise to the latent variable $z$.
We utilize the DDIM sampler~\cite{song2022ddim} to accelerate sampling after training, using 200 sampling steps.
We sampled 100,000 moiré patterns using a single NVIDIA A40 GPU, and some of the samples are shown in Figure~\ref{fig:generated_data}.



\subsection{Moiré Image Synthesis}
In this section, we will demonstrate the implementation details of the Moiré Image Synthesis stage that were omitted in our main paper. Additionally, we include more visualizations of the synthesis results in Figure~\ref{fig:synthesis_result_sup}.

\subsubsection{Implementations of the Moiré Image Blending} 
For the MIB module, $\omega_m$ in Eq. (5) is randomly selected from [0.65, 0.75], while $\omega_g = 1 - \omega_m$. The $op_m$ and $op_g$ in Eq. (6) are set to 1.0 and 0.8, respectively.
Performance changes resulting from the use of both the Multiply and Grain Merge strategy are detailed in the additional ablation study in Section~\ref{sec:sup_ablation}.

\subsubsection{Implementations of the Tone Refinement Network} 
We implement the backbone of our Tone Refinement Network(TRN) using Uformer-T(Tiny)~\cite{Wang2022Uformer}, where the Transformer Block uses the Locally-enhanced Window (LeWin) Transformer block proposed by Uformer and sets the window size to 8$\times$8. At the same time, we change the encoder depth from \{2,2,2,2\} to \{1,1,1,1\}. 
Performance changes resulting from the use of the Uformer are detailed in the additional ablation study in Section~\ref{sec:sup_ablation}.

In the context of TRN, utilizing the transposed convolutional upsampling block similar to Uformer may lead to the emergence of ``Checkerboard Artifacts'' in the output $I_{trn}$, as illustrated in Figure~\ref{fig:checkerboard_artifacts}.
This issue stems from the uneven overlap when transposed convolution is employed during the upsampling process~\cite{odena2016deconvolution}. 
As a solution, we utilize the CARAFE upsampling operator~\cite{wang2019carafe} to replace transposed convolution in Uformer for upsampling. 
CARAFE effectively addresses the "Checkerboard Artifacts" by predicting diverse up-sampling kernels based on the semantic information of the input feature maps~\cite{wang2019carafe}, thereby contributing to improved feature reorganization within TRN.
Performance changes resulting from the use of CARAFE are detailed in the additional ablation study in Section~\ref{sec:sup_ablation}.

We utilize 2 NVIDIA A40 GPUs to train our Tone Refinement Network on UHDM~\cite{yu2022towards} for 50 epochs, FHDMi~\cite{he2020fhde} for 25 epochs, and TIP~\cite{sun2018moire} for 2 epochs. 
The learning rate is initially set to 1e-5 and scheduled by cyclic cosine annealing~\cite{loshchilov2016sgdr}, and models are optimized by Adam~\cite{kingma2014adam} with $\beta_1=\text{0.9}$ and $\beta_2=\text{0.999}$. 
For the input clean natural images, we set the random crop size to 384 $\times$ 384, and for the moiré patterns sampled by the Moiré Pattern Generator, we resized their resolution to 384 $\times$ 384 as well.


\subsubsection{Implementations of loss functions} 
For the loss function Eq. (13) in the main paper, we simply set $\lambda_{per}=\lambda_{color}=\text{1.0}$ and $\lambda_{tv}=\text{0.1}$ to balance the scale of the values during training.

For the perception loss $L_{per}$, to further validate our assumption in the main paper whether computing the content loss between $I_{trn}$ and $I_{mib}$ directly in the pixel space is less effective, we utilize the $\mathcal{L}_1$ loss instead of the $\mathcal{L}_{per}$ loss for our synthesis network in the additional ablation study in Section~\ref{sec:sup_ablation}.

For the color loss $\mathcal{L}_{color}$, we convert $I_{trn}$ and $I_{rm}$ into RGB-uv histogram feature $H(I_{trn})$ and $H(I_{rm})$ from the log-chrominance space followed by prior work on color constancy~\cite{afifi2019sensor, Afifi2019CVPR}, which represents the color distribution of those two images. 
In particular, $u$ and $v$ are used to control the contribution of each color channel in the generated histogram and the smoothness of the histogram bin.
Specifically, given an RGB image $I(\mathrm{x})$ where $\mathrm{x}$ denotes the pixel point index, we first convert it to YUV color space:
\begin{equation}
I_{y}(\mathrm{x})=\sqrt{I_{r}^2(\mathrm{x})+I_{g}^2(\mathrm{x})+I_{b}^2(\mathrm{x})}.
\end{equation}
and:
\begin{flalign}
I_{ur}(i) = \log \frac{I_r(i) + \epsilon}{I_g(i) + \epsilon}  
&\text{;  } I_{vr}(i) = \log \frac{I_r(i) + \epsilon}{I_b(i) + \epsilon} \\
I_{ug}(i) = \log \frac{I_g(i) + \epsilon}{I_r(i) + \epsilon} 
&\text{;  } I_{vg}(i) = \log \frac{I_g(i) + \epsilon}{I_b(i) + \epsilon} \\
I_{ub}(i) = \log \frac{I_b(i) + \epsilon}{I_r(i) + \epsilon} 
&\text{;  } I_{vb}(i) = \log \frac{I_b(i) + \epsilon}{I_g(i) + \epsilon} 
\end{flalign}

where ``$I_r$'', ``$I_g$'', and ``$I_b$'' subscripts refer to the color channels of the image $I$, $\epsilon=\text{10}^{-\text{6}}$ is a small constant added for numerical stability, and ($I_{ur}$, $I_{vr}$), ($I_{ug}$, $I_{vg}$) and ($I_{ub}$, $I_{vb}$) are the $uv$ coordinates of the $I_r$, $I_g$, and $I_b$.

We then generated the unnormalized histogram $H(u,v,c)$ of each color channel $c \in \{r,g,b\}$ according to the HistoGAN~\cite{Afifi2021histogan}, computed as follows:
\begin{equation}
H(u,v,c) \propto \sum_{\mathrm{x}} k\left(I_{uc}(\mathrm{x}), I_{vc}(\mathrm{x}), u, v\right) I_y(\mathrm{x}),
\end{equation}
where $k(\cdot)$ is the inverse-quadratic kernel:
\begin{equation}
\begin{aligned}
k\left(I_{uc}, I_{vc}, u, v\right)=(1 & \left.+\left(\left|I_{uc}-u\right| / \tau\right)^2\right)^{-1} \\
& \times\left(1+\left(\left|I_{vc}-v\right| / \tau\right)^2\right)^{-1}
\end{aligned}
\end{equation}
where $\tau$ is a fall-off parameter to control the smoothness of the histogram's bins. Finally, the histogram features $H(I) \in R^{h \times h \times 3}$ stacked by $H(u,v,c)$ of 3 color channels is normalized to sum to one:
\begin{equation}
H(I) = \frac{[H(u,v,r), H(u,v,g), H(u,v,b)]}{\sum_{u,v,c} H(u, v, c)}.
\end{equation}

Following HistoGAN~\cite{Afifi2021histogan}, we set the histogram bin, $h$, to 64 and set the fall-off parameter of our histogram's bins, $\tau$, to 0.02.
\section{Experiments}
\label{sec:sup_esperiments}
In this section, we will provide a more detailed overview of the experimental setups, present additional visualization results and runtime comparisons, carry out further ablation studies, and address the limitations of our proposed method.

\begin{figure}[!t]
  \centering
    \includegraphics[width=1.0\linewidth]{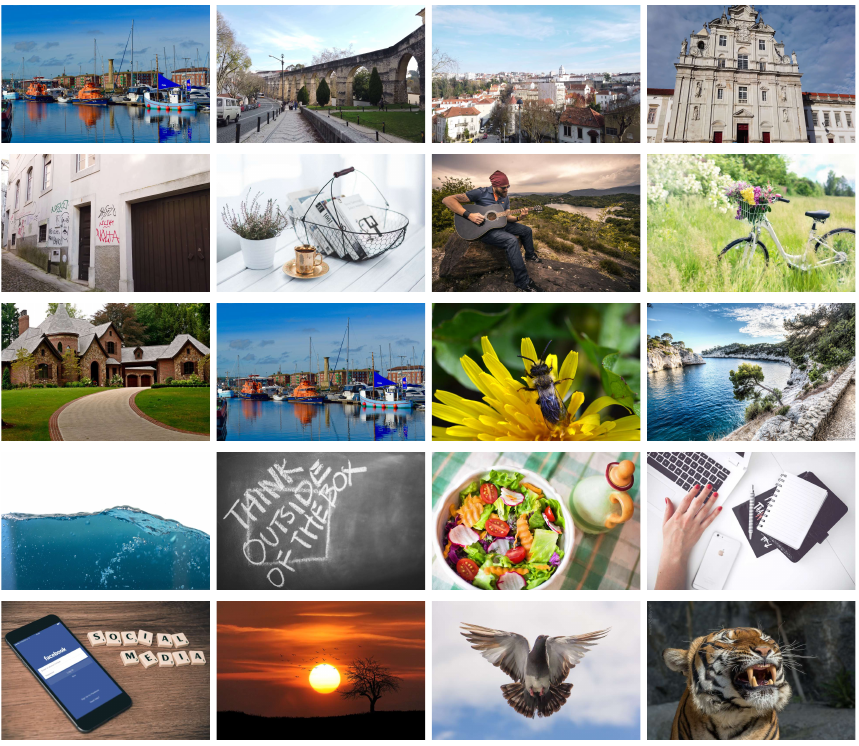}
  \caption{Examples of the MHRNID dataset.} 
  \label{fig:mhrnid}
\end{figure}

\subsection{Experimental Setups}
We implement all the experiments using PyTorch Lightning on multiple NVIDIA A40 GPUs. All experiments were conducted once after setting the seed to the same values as~\cite{yu2022towards} and~\cite{undem}.

\subsubsection{Implementation Details of other Comparison Methods}
For Shooting, we migrated their implementation code from Opencv to PyTorch based on the implementation idea provided by~\cite{shooting}. Note that the Shooting method produces a distorted composite image after random projective transformation. We maintain the transformation parameter and adjust the clean image accordingly to ensure that the moiré image aligns with the clean image during the subsequent demoiréing stage.
For UnDeM~\cite{undem}, we directly use their 384$\times$384 moiré image synthesis network trained on UHDM~\cite{yu2022towards} and FHDMi~\cite{he2020fhde} and also train their synthesis network on TIP~\cite{sun2018moire} in their code framework~\cite{undem}.
For MoireSpace~\cite{yang2023doing}, we utilize the moiré patterns provided by their dataset to obtain the synthesis result by deploying their multiply blending strategy. We resize their moiré patterns to 384$\times$384 for a fair comparison.

\begin{table}[t]
\centering
\setlength{\tabcolsep}{3.5mm}
\begin{tabular}{lcc}
\toprule
Methods          & Params(M) & Runtime(s) \\
\midrule
Shooting         & -         & 0.03       \\
UnDeM            & 2.4       & 0.56       \\
MoireSpace       & -         & 0.01       \\
Ours(MIB)        & -         & 0.02       \\
Ours(MIB \& TRN) & 3.0       & 0.04       \\ \bottomrule
\end{tabular}
\caption{Runtime comparisons.}
\label{tab:Runtime}
\end{table}



\subsubsection{Mixed High-Resolution Natural Image Dataset}
In the Zero-Shot experiments, we collected a comprehensive Mixed High-Resolution Natural Image Dataset (MHRNID) to avoid data overlap between the training and test sets. The MHRNID dataset consists of the super-resolution datasets DF2K-OST~\cite{wang2021real}, the natural image datasets UHD-LOL4K~\cite{wang2023uhdlol4k}, and UHD-IQA~\cite{hosu2024uhdiqa} collated and incorporated, which contains 26,000 high-definition images. We also provide several visual examples of MHRNID, as shown in Figure~\ref{fig:mhrnid}.

\subsubsection{Implementation Details of Demoiréing Models}
For MBCNN~\cite{zheng2020image} and ESDNet-L~\cite{yu2022towards}, we followed the experimental settings from~\cite{yu2022towards} and~\cite{undem}. We trained for 150 epochs on UHDM~\cite{yu2022towards} and FHDMi~\cite{he2020fhde} and 70 epochs on TIP~\cite{sun2018moire}. Additionally, we trained for 50 epochs on the MHRNID dataset.


\subsection{More Qualitative Comparisons}

\subsubsection{Moiré Image Synthesis}
\begin{figure*}[!p]
  \centering
    \includegraphics[width=1\linewidth]{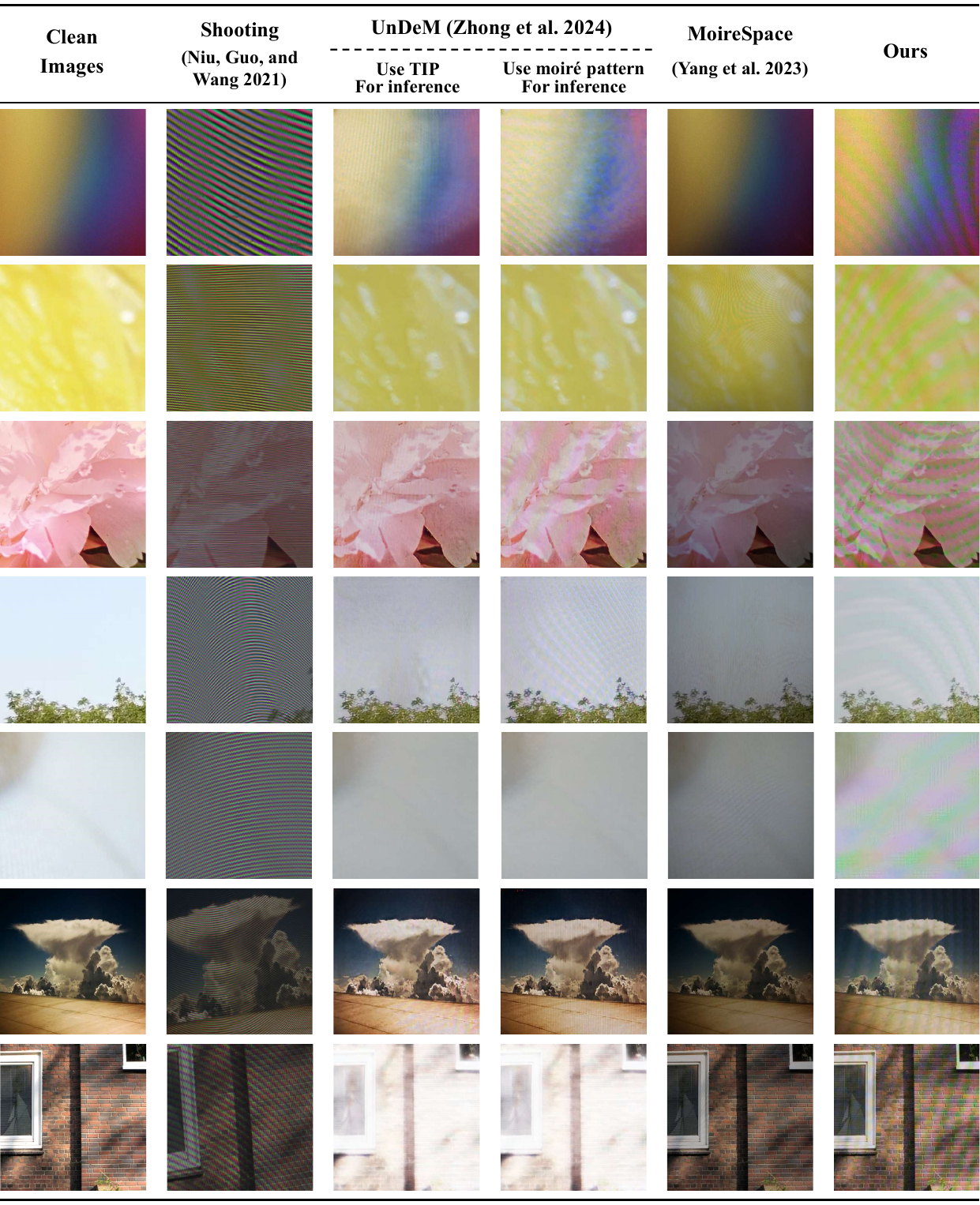}
  \caption{
  Qualitative comparisons of synthesized moire images were obtained using the shooting method, UnDeM, MoireSpace, and our UniDemoiré.
  }
  \label{fig:synthesis_compare}
\end{figure*}

The visualization results of synthesis moiré images on the MHRNID dataset using Shooting~\cite{shooting}, UnDeM~\cite{undem}, and our UniDemoiré are shown in Figure~\ref{fig:synthesis_compare}. 
The moiré image produced by our UniDemoiré is notably superior to other synthesis methods in terms of diversity and realism. In comparison, the moiré image generated by the Shooting~\cite{shooting} method is excessively distorted, UnDeM's network~\cite{undem} is susceptible to anomalies during image generation, and the moiré pattern dataset provided by MoireSpace~\cite{yang2023doing} is of subpar quality. Additionally, the multiplication strategy results in a darker synthesized image.

\subsubsection{Demoiréing}
\begin{figure*}[!p]
  \centering
    \includegraphics[width=1.0\linewidth]{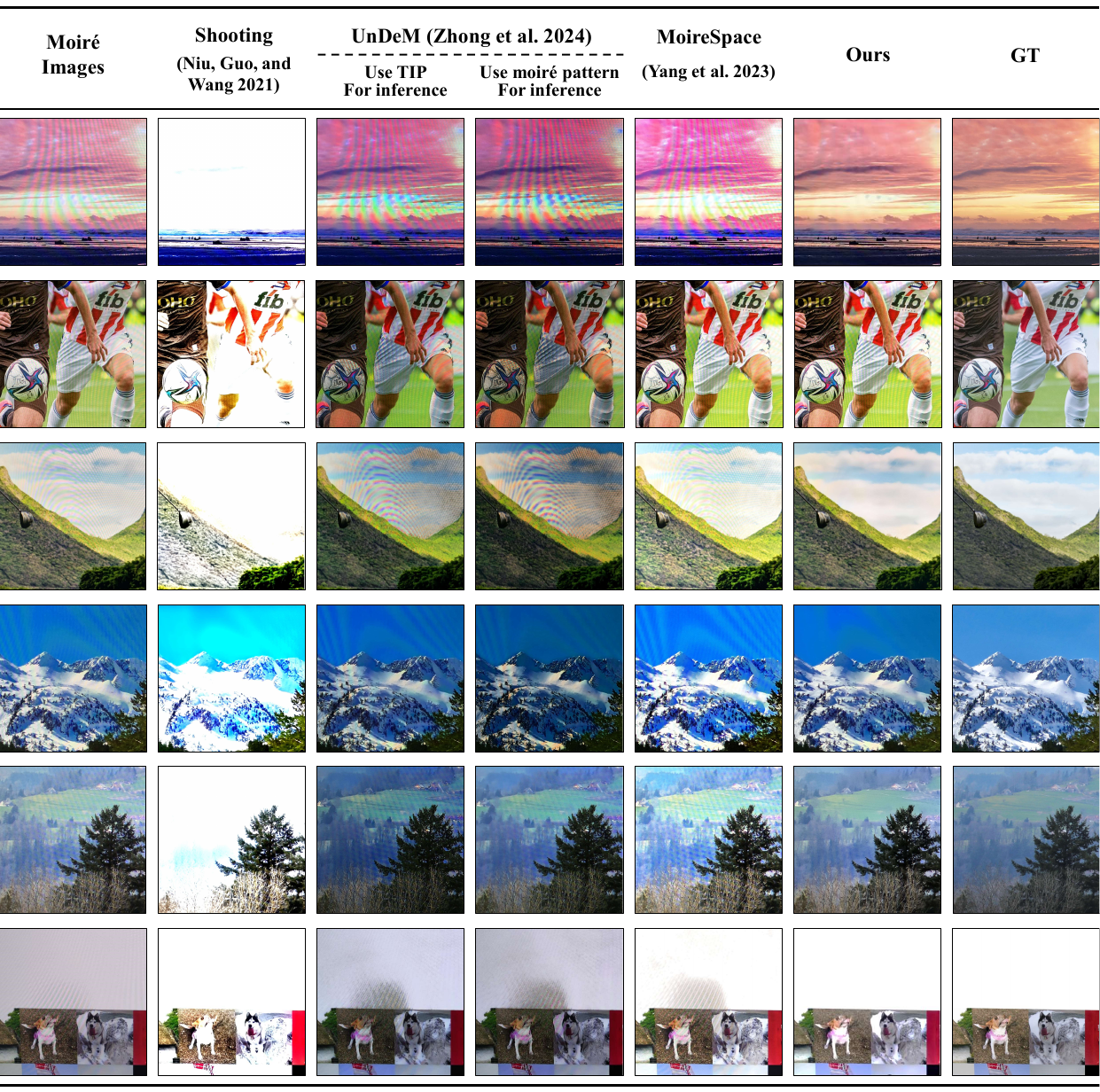}
  \caption{ Qualitative comparisons of zero-shot evaluation on the UHDM dataset.} 
  \label{fig:zero-shot}
\end{figure*}

\begin{figure*}[!p]
  \centering
    \includegraphics[width=1\linewidth]{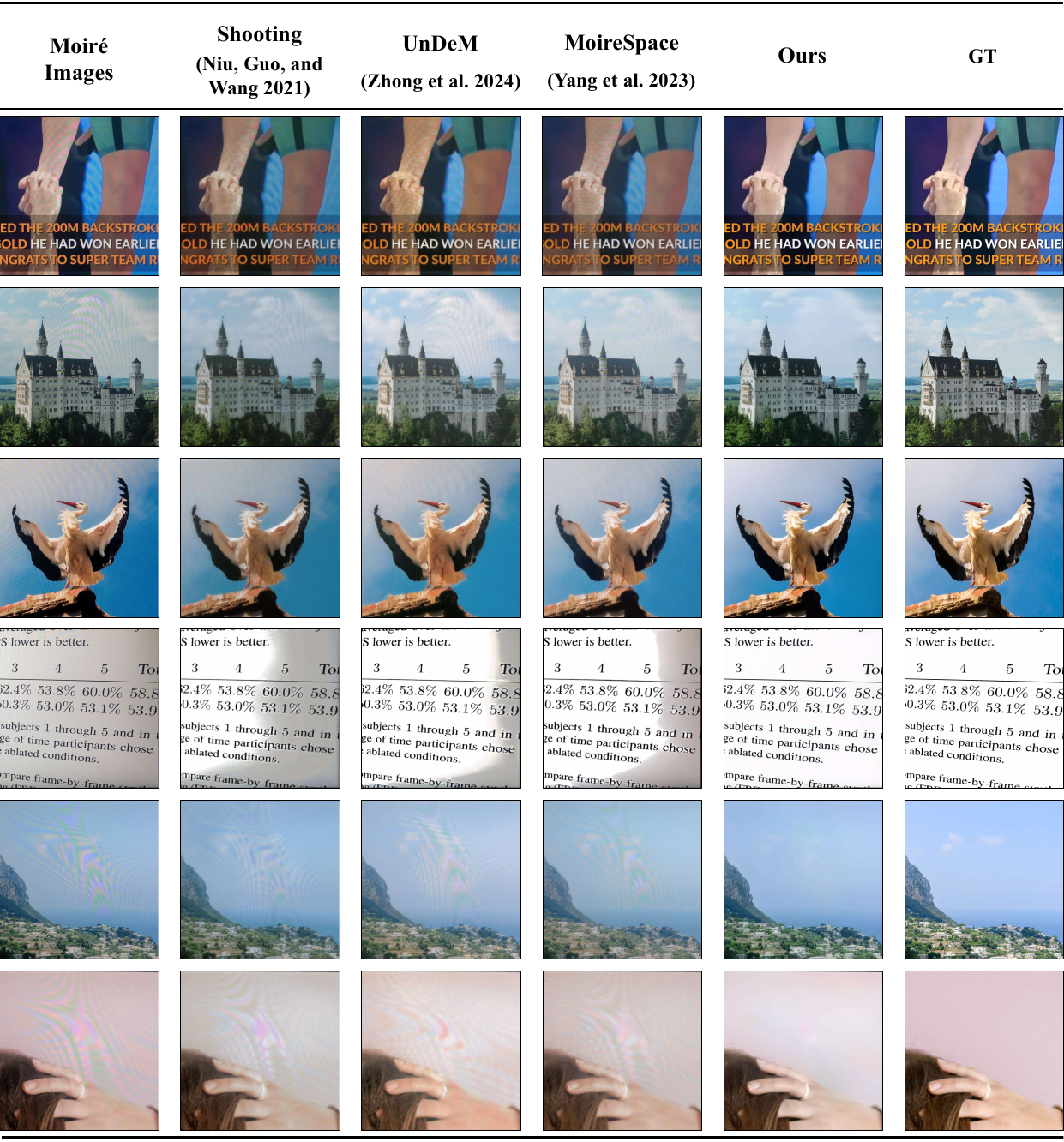}
  \caption{
  Qualitative comparisons of our models with other state-of-the-art methods on the FHDMi dataset.
  }
  \label{fig:cd_fhdmi}
\end{figure*}

\begin{figure*}[!p]
  \centering
    \includegraphics[width=1\linewidth]{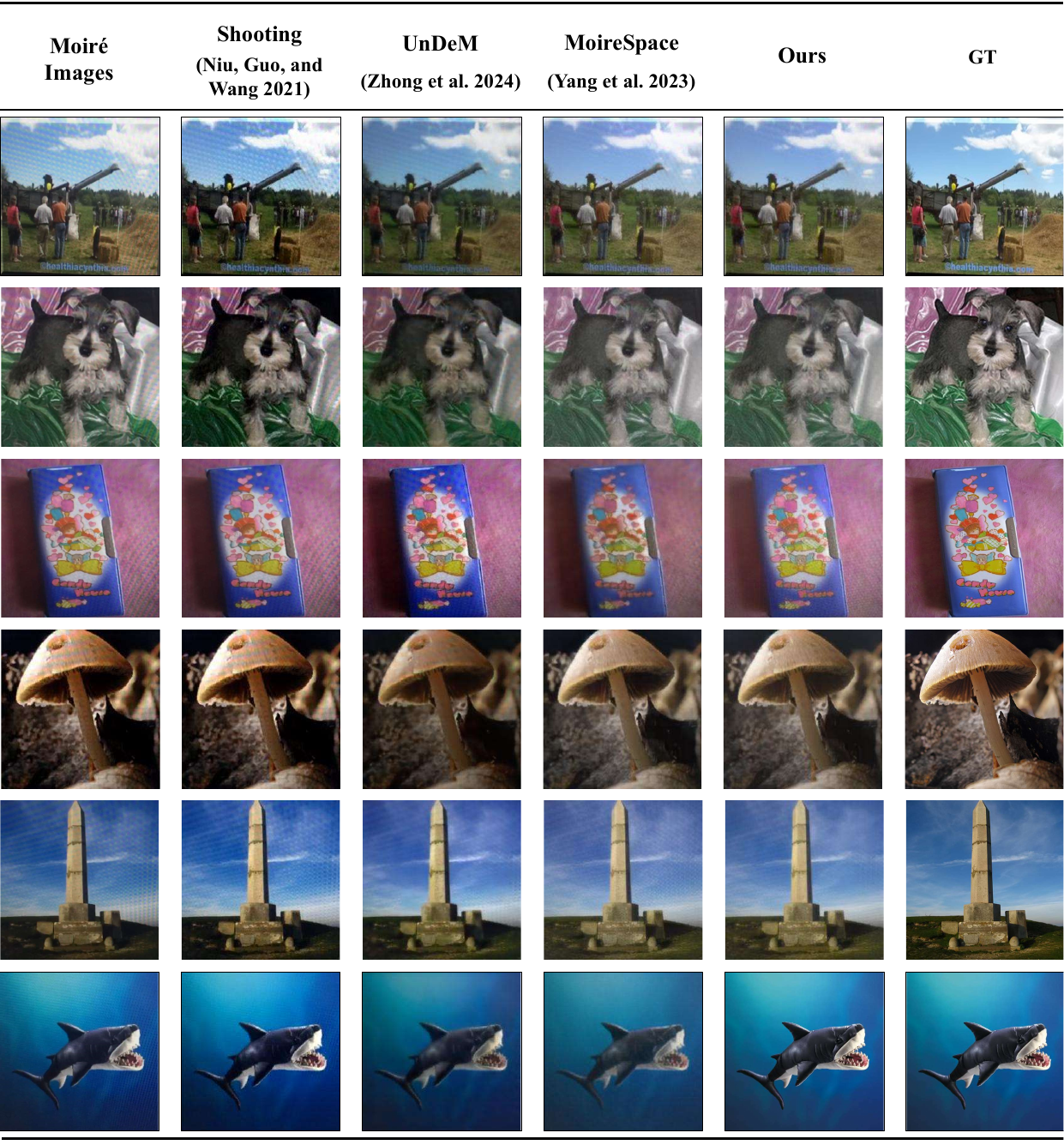}
  \caption{
  Qualitative comparisons of our models with other state-of-the-art methods on the TIP dataset.
  }
  \label{fig:cd_tip}
\end{figure*}

Figure~\ref{fig:zero-shot} shows the visualization results of zero-shot demoiréing on UHDM~\cite{yu2022towards}. Additionally, Figures~\ref{fig:cd_fhdmi} and~\ref{fig:cd_tip} illustrate the demoiréing results on FHDMi~\cite{he2020fhde} and TIP~\cite{sun2018moire} using ESDNet-L~\cite{yu2022towards} trained on UHDM~\cite{yu2022towards}. Our method's model effectively removes moiré artifacts and retains high-frequency details, indicating the strong generalization ability of our proposed UniDemoiré.



\begin{figure}[t]
  \centering
    \includegraphics[width=1.0\linewidth]{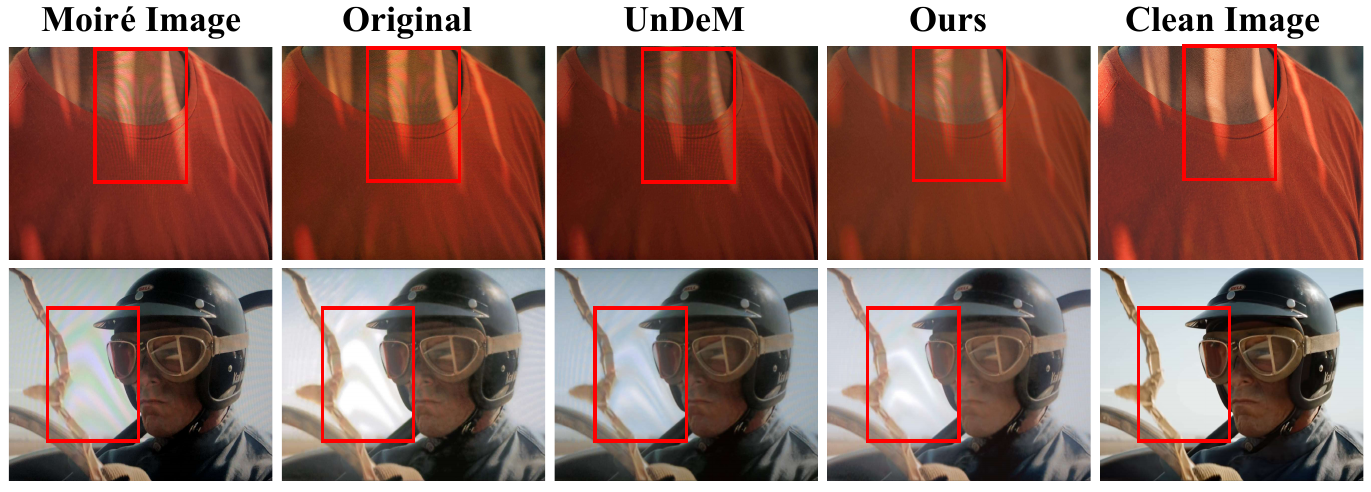}
\vspace{-1ex}
 \caption{Failure Examples.} 
  \label{fig:limitations}
\end{figure}

\subsection{Runtime Comparisons}
\label{sec:runtime}
Table~\ref{tab:Runtime} shows the comparison of the parameters and the running time of our synthesis module with other methods. 
To ensure fair comparisons, our method and UnDeM use torchinfo for parameter counting, with all methods utilizing 256x256 input images.
Our experimental results indicate that our method, slightly exceeding UnDeM in parameters, achieves a runtime comparable to non-learning algorithms like Shooting and MoireSpace, demonstrating the efficiency of our MIB and TRN.
Furthermore, our model's FLOPs are 5.6266G, significantly lower than UnDeM's 26.7576G, indicating high performance and reduced computational cost.

\subsection{Additional Ablation Study}
\label{sec:sup_ablation}
The results of the additional ablation experiments are in Table~\ref{tab:Exp_ablation_sup}. 
where ``$\mathcal{L}_{per} \rightarrow \mathcal{L}_1$'' denotes replacing the perception loss $\mathcal{L}_{per}$ in the synthesis network with the L1 loss $\mathcal{L}_1$. ``Uformer $\rightarrow$ UNet" denotes switching the entire backbone network of the TRN from Uformer to UNet~\cite{ronneberger2015u}. For a fair comparison, we kept the number of upsampling/downsampling blocks and the base channel in UNet consistent with Uformer, while removing the attention block.

\begin{table}[h]
\centering
\setlength{\tabcolsep}{1.9mm}
\scalebox{1.0}{
\begin{tabular}{lccc}
\toprule
Components                     &\ua{PSNR}   &\ua{SSIM} &\da{LPIPS} \\ 
\midrule
ALL                                                      & \textbf{20.7543} & \textbf{0.7653} & \textbf{0.2136} \\
MIB ($w/o$ Multiply)                                     & 20.3158          & 0.7598          & 0.2328          \\
MIB ($w/o$ Grain Merge)                                  & 20.3930          & 0.7587          & 0.2414          \\
TRN ($w/o$ CARAFE)                                       & 20.4414          & 0.7408          & 0.2256          \\
TRN ($\mathcal{L}_{per}$ $\rightarrow$ $\mathcal{L}_1$)  & 20.1404          & 0.7447          & 0.2495          \\
TRN (Uformer $\rightarrow$ UNet)                         & 20.3899          & 0.7476          & 0.2413          \\
\bottomrule
\end{tabular}
}
\caption{Additional ablation studies. Source: UHDM, Target: FHDMi.}

\label{tab:Exp_ablation_sup}
\end{table}

The results of two sets of ablation experiments on layer blending strategies also show that using only one of them leads to distortion of the synthesis results, which in turn affects the model's generalization ability.
The results of the ``$\mathcal{L}_{per}\rightarrow\mathcal{L}_1$'' show that computing the loss function in this way leads to a degradation of the model performance because moiré patterns can disrupt image structures by generating strip-shaped artifacts. 
The results of the ``$w/o$ CARAFE'' indicate that using the CARAFE upsampling operator~\cite{wang2019carafe} yields better fusion performance than the transposed convolution originally employed by Uformer~\cite{Wang2022Uformer}.
Furthermore, the results from the “Uformer $\rightarrow$ UNet” demonstrate that the LeWin Transformer Block within Uformer is more effective at extracting color features from moiré patterns compared to the original UNet architecture.


\subsection{Limitations}

In some cases, particularly when the moiré artifacts in the target domain significantly differ from those in the source domain, our solution may struggle to completely remove all artifacts, as Figure~\ref{fig:limitations} shows. However, even in these challenging scenarios, our method tends to perform better at artifact removal compared to the baselines. Our performance can be further refined by generating more diverse moiré patterns and synthesized training data.
In Figure~\ref{fig:limitations}, we show a failure case. When the moiré artifacts in the target domain are too different from those in the source domain, our solution still struggles to produce a completely moiré-free result. However, we still remove the artifacts comparatively better than baselines.

\newpage
\clearpage

\end{document}